\documentclass[journal]{IEEEtran}
\usepackage{newtxtext}
\usepackage{amsmath,amssymb,amsfonts}
\usepackage{graphicx}
\usepackage{booktabs}
\usepackage{multirow}
\usepackage{tabularx}
\usepackage{array}
\usepackage{cite}
\usepackage{comment}
\makeatletter
\let\NAT@parse\undefined
\makeatother
\usepackage{hyperref}
\usepackage{relsize}
\usepackage{float}
\usepackage{pifont}
\usepackage[dvipsnames]{xcolor}
\usepackage{url}
\usepackage{hyperref}
\usepackage{cleveref}

\definecolor{gain}{RGB}{0,128,0}

\title{M3GD: Multi-Modal Multi-View Geometric Diffusion for Camera--LiDAR Novel View Synthesis}
\author{Yang Zhou$^{1,*}$, Jiuhong Xiao$^{1,2,*}$, Shizhao Ye$^{2}$, Long Quang$^{1,3}$, Carlos Nieto-Granda$^{3}$, and Giuseppe Loianno$^{2}$
\thanks{$^{*}$These authors contributed equally to this work.}
\thanks{$^{1}$The authors are with New York University, New York, USA. {\tt\footnotesize email: \{yangzhou, jx1190, lq2146\}@nyu.edu}.}
\thanks{$^{2}$The authors are with Department of Electrical Engineering and Computer Sciences, University of California, Berkeley, CA 94720, USA. {\tt\footnotesize email: \{jiuhong.xiao, yeshizhao, loiannog\}@berkeley.edu}.}
\thanks{$^{3}$The authors are with the U.S. Army Combat Capabilities Development Command, Army Research Laboratory, Adelphi, MD 20783, USA. {\tt\footnotesize email: \{long.p.quang.civ, carlos.p.nieto2.civ\}@army.mil}.}
}

\begin{document}
\maketitle

\begin{abstract}
Robotic novel view synthesis (NVS) must recover both visual appearance and metric 3D structure, yet most generative NVS methods rely only on images, overlooking LiDAR, a complementary sensor common on robotic platforms. We present M3GD, a Camera--LiDAR multimodal representation for generative NVS that composes independently pretrained 2D image and 3D point-cloud foundation models without separately pretraining a cross-modal translator.
We show that, after camera projection, frozen LiDAR and image features exhibit substantial shared spatial structure, providing a natural cross-modal representation.
M3GD conditions generation on LiDAR through this structure: it combines explicit geometry statistics with learned point-cloud descriptors into view-aligned packets on the image-latent grid, injected through a lightweight residual adapter into a multi-view flow-matching generator whose latent space, decoders, and training objective remain intact.
On the GrandTour dataset, M3GD improves target-view RGB and depth synthesis over an image-only version of the same backbone.
Ablations show that the gains come from pixel-aligned LiDAR content and that target-view LiDAR acts as a geometric query linking the requested view to source observations.
Deployment on a ground robot demonstrates practical real-world operation, with a configurable quality--cost trade-off controlled by the number of Euler integration steps.
\end{abstract}

\section{Introduction}\label{sec:introduction}
Novel view synthesis (NVS), i.e., predicting the image a camera would capture from an unvisited viewpoint, turns recorded or mapped environments into reusable visual testbeds for robots. View-controllable rendering has been used to train or evaluate learning-based driving and drone policies in closed-loop simulation~\cite{amini2022vista,yang2023unisim,chen2026gradnavpp}. The underlying radiance-field representations also support photorealistic mapping for localization and geometric reconstruction~\cite{pan2025pings}, while synthesized viewpoints give teleoperators visual coverage beyond the physical camera layout of a robot~\cite{wildersmith2024rfteleoperation}. Recent generative NVS systems have advanced rapidly by building on 2D image representations and multi-view diffusion models~\cite{gao2024cat3d,guizilini2025mvgd,jang2026gld}. However, computer vision benchmarks usually pose NVS as RGB-to-RGB synthesis from posed images alone. Conversely, robotic deployment changes the input boundary: cameras are commonly synchronized with calibrated LiDAR, metric structure is operationally important, and a LiDAR sweep near the queried pose may exist even when the target RGB image is unavailable. We treat such target-side LiDAR as a \emph{geometric query}: it anchors the requested viewpoint and local metric support without revealing the target appearance to be synthesized. This matters especially where RGB-only synthesis is difficult: texture-poor corridors, thin structures, harsh illumination, and long-range outdoor geometry. The core question is therefore not whether another sensor can be added, but whether pretrained 2D and 3D representations can be composed so that visual appearance and metric structure help the same generative NVS model.

Most existing NVS methods address only part of this problem. Generative approaches synthesize plausible target views from RGB observations and their associated camera poses~\cite{liu2023zero1to3,liu2024syncdreamer,sargent2024zeronvs,chan2023genvs}, while feed-forward radiance-field and splatting methods reconstruct renderable scene representations from sparse images~\cite{yu2021pixelnerf,gao2024cat3d,charatan2024pixelsplat,chen2024mvsplat,xu2025depthsplat}. A separate line of work employs depth or LiDAR to supervise view synthesis, but typically requires fitting or updating an object- or environment-specific representation before novel views can be queried~\cite{deng2022dsnerf,rematas2022urban,yang2023unisim,tonderski2024neurad}. Together, these limitations motivate the RGB--LiDAR-conditioned generative NVS setting studied in this work. Given posed source RGB--LiDAR observations and a target query comprising a camera pose and synchronized LiDAR sweep, the goal is to predict the target-view RGB image and dense depth without fitting a scene-specific representation at deployment.

The challenge in combining these sensing modalities stems from their heterogeneity. Images are dense, appearance-rich, and indexed by pixels whereas LiDAR data is sparse, unordered, and indexed by metric 3D coordinates. Image evidence carries texture, geometry, illumination, and semantic context, whereas LiDAR carries scale-accurate support, range, and free-space structure. Their learned representations also differ in spatial support, dimensionality, pretraining objective, and invariances. Naively, collapsing LiDAR to a dense depth map discards much of its support structure, while forcing a generative model to learn a full cross-modal translator turns fusion into an additional synthesis problem.

This paper studies RGB--LiDAR fusion as a representation-composition problem. Collecting synchronized, calibrated multimodal robot data across diverse environments limits the scale available for training each NVS system from scratch, making it attractive to reuse visual and geometric priors learned from broader pretraining data. Rather than training image--point alignment explicitly~\cite{sautier2022slidr,peng2023openscene,chen2023clip2scene} or rebuilding the NVS model around a new sensor stack, we ask whether a pair of independently trained 2D image and 3D point-cloud foundation representations is compatible after camera projection. If a projected 3D representation is legible to the image-latent space of a given generator, then LiDAR can enter NVS as spatially aligned metric evidence instead of being reduced to an auxiliary loss, a per-scene reconstruction prior, or a single hand-crafted depth image.

We instantiate this perspective with M3GD, an RGB--LiDAR multimodal representation for generative NVS. We analyze a particular pairing of frozen Depth Anything 3 (DA3) image features and Utonia point-cloud features~\cite{lin2025depthanything3,zhang2026utonia} and find that they share substantial spatial structure after camera projection. Each M3GD LiDAR packet then combines two complementary representations derived from the same sweep on the same image-latent grid: explicit geometry statistics retain direct information about metric range, sampling support, and sensor timing, while pooled Utonia descriptors expose learned 3D structure from point-cloud pretraining. Their concatenation provides a view-aligned representation between LiDAR and the generative NVS backbone. The resulting design keeps the paper's focus: how to combine dense appearance reasoning with both explicit and learned sparse 3D evidence without separately pretraining a cross-modal translator.

Our contributions are therefore summarized as follows:
\begin{itemize}
\item We identify a specific class of independently trained, frozen 2D image and 3D point-cloud foundation representation pairs whose projected 3D features share substantial spatial structure with the 2D image latents. This cross-modal compatibility motivates their incorporation into an image-latent generative model.
\item We introduce M3GD, an RGB--LiDAR multimodal conditioning method for generative NVS that combines explicit geometry statistics with learned point-cloud foundation-model descriptors in a projected feature representation and injects it into an image-latent generative model trained with multimodal flow matching, without requiring separate pretraining of a cross-modal translator.
\item We evaluate M3GD on GrandTour and show that adding projected LiDAR packets improves RGB and depth synthesis over an image-only version.
\item We deploy M3GD on a wheeled ground robot with a different camera--LiDAR sensor suite in a real-world environment without any retraining on the target domain. This demonstrates that the learned conditioning generalizes across platforms, and establishes its practical relevance for robotic perception tasks.
\end{itemize}
 
\section{Related Works}\label{sec:related-work}
\subsection{Generative Novel View Synthesis in Geometry-Aware Latents}\label{subsec:generative-novel-view-synthesis-in-geometry-aware-latents}

Diffusion-based NVS initially matured around camera-conditioned image diffusion. Zero-1-to-3~\cite{liu2023zero1to3} demonstrated that large pretrained 2D diffusion models encode viewpoint priors steerable by relative camera pose. SyncDreamer~\cite{liu2024syncdreamer} improved multi-view consistency by denoising the joint distribution of target views. ZeroNVS~\cite{sargent2024zeronvs} extended camera-conditioned priors from objects to full scenes. A second family couples generation with reconstruction: GeNVS~\cite{chan2023genvs} denoises through a latent 3D feature field, ReconFusion~\cite{wu2024reconfusion} regularizes per-scene NeRF optimization with a novel-view diffusion prior, and CAT3D~\cite{gao2024cat3d} generates additional views for a downstream reconstruction pipeline. Although these methods introduce camera geometry, depth prediction, or internal 3D representations, their scene input remains a single modality: RGB images and their associated poses. They establish that generation suits target-view ambiguity, but provide no conditioning pathway for measurements or learned features from a physical 3D sensor.

Closer to our work, some methods move generation into a geometry-aware space. MVGD~\cite{guizilini2025mvgd} jointly generates target-view RGB and scale-consistent depth from images and corresponding poses via ray-conditioned multi-view geometric diffusion. GLD~\cite{jang2026gld} goes further, diffusing directly in the frozen feature space of a geometric foundation model (DA3~\cite{lin2025depthanything3} or VGGT~\cite{wang2025vggt}) with a DiT-style transformer~\cite{peebles2023dit,zheng2025rae} trained under flow matching~\cite{lipman2023flow,ma2024sit}. This latent space already encodes cross-view correspondence, and frozen decoders recover RGB and depth from synthesized latents. Crucially, GLD is an \emph{image-input-only} baseline: it conditions on posed source RGB images and consumes no LiDAR or other measured 3D point cloud. Therefore, its geometric awareness comes from image-derived features rather than direct 3D sensor observations. M3GD retains GLD's generative backbone but extends its conditioning from unimodal to multimodal, taking source RGB images together with projected features from measured 3D LiDAR point clouds. Our contribution is thus not another single-modality geometry-aware latent but a cross-modal conditioning method that lets 2D appearance and 3D measurements jointly condition latent-space generative NVS without separately pretraining a cross-modal translator.

\subsection{Generalizable Feed-Forward View Synthesis}\label{subsec:generalizable-feed-forward-view-synthesis}

Deterministic feed-forward methods are the strongest non-generative baseline class for sparse-view synthesis. PixelNeRF~\cite{yu2021pixelnerf} and CAT3D~\cite{gao2024cat3d} condition NeRFs~\cite{mildenhall2020nerf} on projected source-view features. MVSNeRF~\cite{chen2021mvsnerf} and GeoNeRF~\cite{johari2022geonerf} strengthen geometric reasoning with plane-swept cost volumes. The explicit branch predicts renderable primitives directly: pixelSplat~\cite{charatan2024pixelsplat} regresses 3D Gaussians~\cite{kerbl20233dgs} from image pairs, MVSplat~\cite{chen2024mvsplat} anchors Gaussian centers with a cost-volume prior, and DepthSplat~\cite{xu2025depthsplat} couples feed-forward splatting with a pretrained monocular depth backbone. Query-based reconstruction transformers~\cite{chen2026nova3r,peng2026uniquer} drop pixel alignment altogether. Trained on large multi-view datasets such as RealEstate10K~\cite{zhou2018realestate10k} and DL3DV~\cite{ling2024dl3dv}, all these feed-forward models derive geometry solely from images, treating vision as the single input modality. None of the approaches consumes a pretrained point-cloud representation at inference time, and none produces both RGB and depth from a shared pretrained image latent. We compare against MVSplat and DepthSplat as representative deterministic competitors (Section~\ref{sec:experiments}). In contrast, M3GD uses conditional diffusion in a shared image-latent space: rather than mapping RGB observations to a single renderable scene estimate, it iteratively denoises a target latent conditioned on both images and projected LiDAR foundation features. This generative formulation better suits sparse-view synthesis, where occlusions, disocclusions, and incomplete image geometry leave target appearance underdetermined: a learned visual prior resolves those ambiguities without forcing a single feed-forward regression, and mutually informed RGB and depth are decoded from the synthesized latent.

\subsection{Depth and LiDAR Assisted View Synthesis}\label{subsec:depth-and-lidar-as-side-information-for-view-synthesis}
Robotic view synthesis supports policy evaluation, sensor-data augmentation, and viewpoint-flexible teleoperation, so sensor coverage and turnaround time are both operational concerns. VISTA 2.0~\cite{amini2022vista} synthesizes novel RGB, 3D LiDAR, and event-camera observations from recorded trajectories for closed-loop autonomous-driving experiments. Radiance Fields for Robotic Teleoperation~\cite{wildersmith2024rfteleoperation} instead trains NeRF or Gaussian-splatting representations online from live robot data so operators can inspect viewpoints the physical cameras miss. These systems establish the robotics value of novel views, but also expose a practical distinction: fast rendering from a reconstructed scene does not remove the time needed to build or update the scene representation.

Most depth- and LiDAR-assisted rendering methods use geometry as \emph{supervision for per-scene optimization}. DS-NeRF~\cite{deng2022dsnerf} supervises NeRF with sparse SfM depth, SparseNeRF~\cite{wang2023sparsenerf} distills depth ranking from noisy sensors, and RegNeRF~\cite{niemeyer2022regnerf} and DiffusioNeRF~\cite{wynn2023diffusionerf} regularize unseen-view appearance and geometry. Urban Radiance Fields~\cite{rematas2022urban} adds LiDAR losses for street-level NeRFs. Robotics systems extend this to multimodal reconstruction: NeuSim~\cite{yang2023neusim} fuses camera and LiDAR to reconstruct objects for sensor simulation, SiLVR~\cite{tao2026silvr} uses range constraints to improve large-scale LiDAR--visual NeRF geometry in texture-poor regions, and PINGS~\cite{pan2025pings} couples a LiDAR-derived distance field with a Gaussian-splatting radiance field for incremental robotic mapping. Despite differing representations, these approaches construct, optimize, or update an object- or environment-specific model before novel views can be queried.

Robotics also changes the input boundary: calibrated LiDAR is often a native observation rather than geometry inferred from images. The LiDAR-native NVS approaches such as NFL~\cite{huang2023nfl}, LiDAR-NeRF~\cite{tao2023lidarnerf}, and LiDAR4D~\cite{zheng2024lidar4d} synthesize LiDAR returns and emphasizes that LiDAR carries support structure destroyed when the sensor collapses into a single pseudo-depth map. Camera--LiDAR driving simulators such as UniSim~\cite{yang2023unisim} and NeuRAD~\cite{tonderski2024neurad} fuse both sensors to re-render driving scenes, but at the cost of per-scene optimization: each scene demands extra fitting before it can be queried, and the resulting representation is valid only for that scene. M3GD instead amortizes RGB--LiDAR fusion into a generalizable generative model and keeps LiDAR as a structured conditioning signal: it avoids scene-specific fitting at deployment and moves LiDAR from an optimization loss or output modality into a reusable conditioning signal, available on both source views and target-view geometric queries.

\subsection{Geometric-Aware Foundation Models and Data Alignment}\label{subsec:foundation-models-for-geometry-and-their-alignment}

Collecting a task-specific robotics dataset requires synchronized, calibrated multimodal sensors across representative environments, limiting the scale and diversity available for training each robotic NVS model from scratch. Robotics representation learning therefore increasingly borrows from broader pretraining data. HRP~\cite{srirama2024hrp} distills affordances from internet-scale human video into representations for robot learning. Geometric image foundation models transfer broad visual and geometric priors: DUSt3R~\cite{wang2024dust3r} regresses dense pointmaps, VGGT~\cite{wang2025vggt} jointly predicts cameras, depth, and point tracks from many views, CUT3R~\cite{wang2025cut3r} adds a persistent state for streaming 3D perception, and the Depth Anything series~\cite{yang2024depthanything2,lin2025depthanything3} provides robust transferable depth encoders. MapAnything~\cite{keetha2025mapanything} factorizes the reconstruction interface into rays, depth, pose, and scale, a perspective our asymmetric source/target packet design shares. On the 3D side, self-supervised point transformers~\cite{wu2024ptv3,wu2025sonata} have culminated in Utonia~\cite{zhang2026utonia}, a single point-cloud encoder across domains whose features have already proven useful for embodied reasoning. These 2D and 3D models offer complementary priors learned from far broader data than a single robot collection.

Our alignment study connects these families. Representation-similarity tools such as linear CKA~\cite{kornblith2019cka} and SVCCA~\cite{raghu2017svcca} typically compare networks trained on the same modality. We adress the \emph{cross-modality} challenge by comparing projected Utonia LiDAR features against DA3 and DINOv2~\cite{oquab2024dinov2learningrobustvisual} image features on spatially overlapping grids. To our knowledge, the substantial shared spatial structure we observe between independently trained 2D and 3D foundation encoders (Section~\ref{subsec:alignment-study}), and its use for generative NVS, have not been reported before.

This \emph{emergent} structure differs from works that learn image--point alignment explicitly. Cross-modal distillation methods such as PPKT~\cite{liu2021ppkt}, SLidR~\cite{sautier2022slidr}, Seal~\cite{liu2023seal}, OpenScene~\cite{peng2023openscene}, and CLIP2Scene~\cite{chen2023clip2scene} supervise a 3D encoder to reproduce 2D (often CLIP) features, showing that alignment can be engineered. Robotics methods pursue similar engineered fusion mechanisms: SGR~\cite{zhang2023sgr} combines semantics from a pretrained 2D model with depth-based 3D spatial reasoning, while F3RM~\cite{shen2023f3rm} distills pretrained image features into a scene-specific 3D feature field for manipulation. We study a complementary question: whether two off-the-shelf encoders trained separately on different modalities and broader pretraining corpora already share enough structure to be composed after projection. This is consistent with the Platonic Representation Hypothesis~\cite{huh2024platonic} that models trained on different data and objectives converge toward a shared world representation, and gives it a concrete robotics test: if projected 3D features share spatial structure with the image latent, a robot can borrow complementary 2D and 3D priors through a lightweight learned conditioning pathway rather than pretrain a new aligned representation from limited task data. Relative to both geometric foundation models and explicit image--point distillation, our contribution is twofold: we \emph{measure} zero-shot 2D--3D representation similarity after projection, and turn that observation into an NVS conditioning method instead of separately pretraining an aligned encoder.
 
\section{Problem Statement and Preliminaries}\label{sec:preliminaries}
To ground the projected Camera--LiDAR conditioning used by M3GD, we first state the NVS problem, subsequently introduce the diffusion and flow-matching machinery used by our backbone, and finally summarize the 2D and 3D foundation representations composed by M3GD.

\subsection{Problem Setup}\label{subsec:problem-setup}
We consider $N$ posed source views $\{(I^\mathrm{src}_i, \mathsf{K}^\mathrm{src}_i, T^\mathrm{src}_i)\}_{i=1}^{N}$, where $I^\mathrm{src}_i$ is the $i^\text{th}$ RGB image with camera intrinsics $\mathsf{K}^\mathrm{src}_i$ and camera-to-world pose $T^\mathrm{src}_i$, each paired with a LiDAR observation $P^\mathrm{src}_i$. We also consider $M$ target cameras $\{(\mathsf{K}^\mathrm{tgt}_j, T^\mathrm{tgt}_j)\}_{j=1}^{M}$ for which LiDAR observations $P^\mathrm{tgt}_j$ may be available but RGB images are not. The goal is to predict the target-view RGB images $I^\mathrm{tgt}_j$ and \emph{dense} depth maps $D^\mathrm{tgt}_j$, defined at every pixel. When the role or index is immaterial, we drop the superscript and subscript and write $I$, $P$, etc. When source and target views are processed jointly, we index all $V{=}N{+}M$ views by a single subscript $v$, with $\mathcal{V}^\mathrm{src}$ and $\mathcal{V}^\mathrm{tgt}$ denoting the source and target index sets, and write $I_v$ and $P_v$ for the image and LiDAR sweep of view $v$. $I_v$ is an input only for $v\in\mathcal{V}^\mathrm{src}$. For $v\in\mathcal{V}^\mathrm{tgt}$, it is the quantity to be predicted.
In our robotic setting, synchronized camera and LiDAR streams are recorded continuously, so a LiDAR sweep near the target timestamp can be available even when the corresponding RGB frame is not provided to the model. We treat this sweep as a geometric query: sparse metric evidence about the target view without providing either the held-out RGB image or the dense depth map to be predicted (Section~\ref{subsec:projected-rgb-lidar-representation}).

\subsection{Diffusion and Flow Matching}\label{subsec:diffusion-models}\label{subsec:flow-matching}
\begin{figure}[t]
    \centering
    \includegraphics[width=\columnwidth]{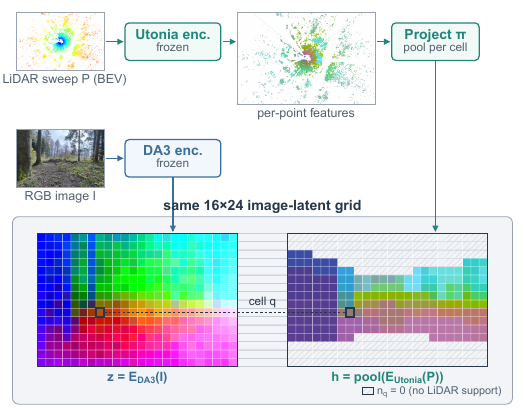}
    \vspace{-10pt}
    \caption{Projected RGB--LiDAR representation. Frozen Utonia per-point descriptors are projected and pooled into the same $H{\times}W$ image-latent grid produced by the frozen DA3 encoder, so each cell $q$ indexes both $z_q$ and $h_q$ (first three PCA components; hatched cells have no LiDAR support, $n_q{=}0$).}
    \label{fig:projected-representation}
    \vspace{-10pt}
\end{figure}
\begin{figure*}[!t]
    \centering
    \includegraphics[width=\textwidth]{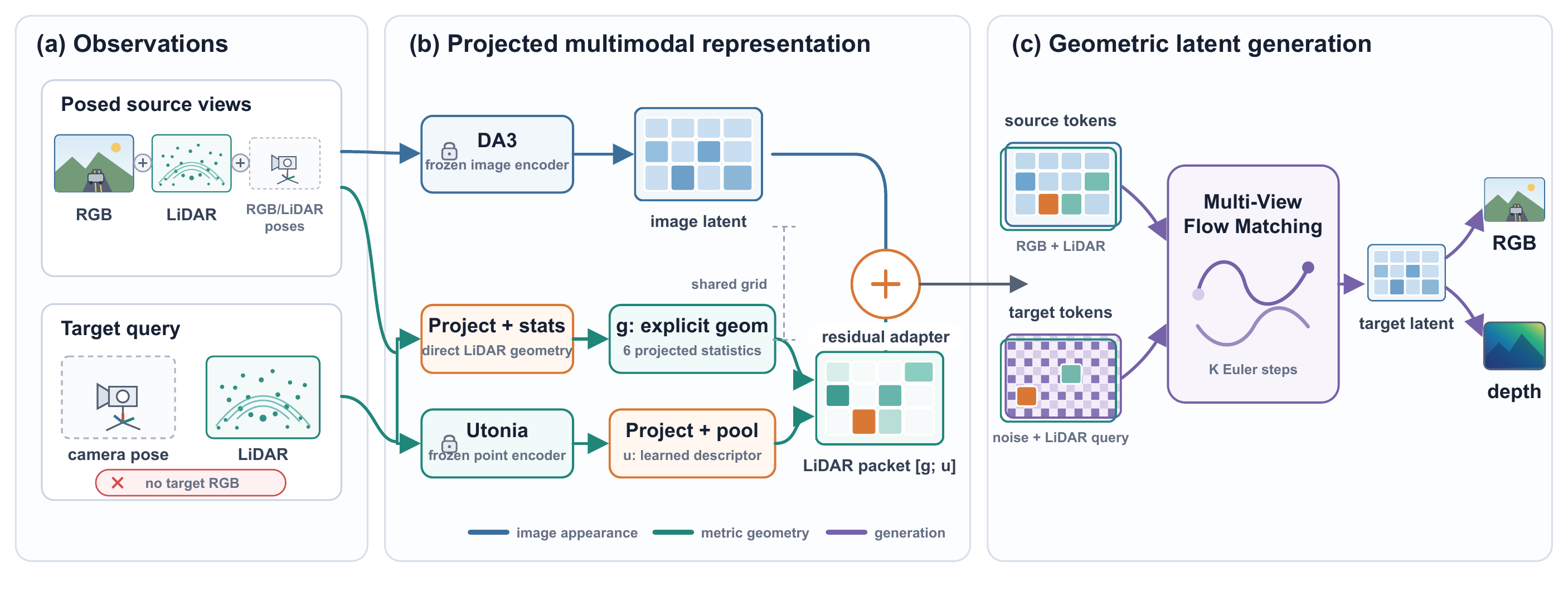}
    \vspace{-30pt}
    \caption{Overview of M3GD. (a) Posed source views provide RGB, LiDAR, and sensor poses, while the target query supplies only a camera pose and LiDAR; target RGB is never observed. (b) The frozen DA3 encoder maps source images to image latents. From the same LiDAR sweep, camera projection constructs explicit geometry statistics, while the frozen Utonia encoder supplies learned per-point descriptors that are pooled into the corresponding image-latent cells; their concatenation forms the per-view LiDAR packet. Fig.~\ref{fig:projected-representation} details this projected RGB--LiDAR representation. A zero-initialized residual adapter composes this packet with the DA3 grid cell by cell. (c) The resulting source tokens (RGB + LiDAR) and target tokens (noise + LiDAR query) enter the multi-view flow matching network, which integrates the learned velocity field for $K$ Euler steps to produce the target latent; frozen decoders recover RGB and depth. Arrow colors distinguish the image-appearance (blue), metric-geometry (teal), and generation (purple) pathways.}
    \label{fig:m3gd-pipeline}
    \vspace{-10pt}
\end{figure*}

A diffusion model is a conditional generator that learns to recover a clean sample from a deliberately corrupted one~\cite{ho2020ddpm}: a network is trained to predict a quantity pointing back toward the clean data distribution, such as the injected noise, the clean sample, or an equivalent velocity. Let $z_0$ denote the representation to be generated and $y$ the conditioning information. In pixel-space image generation, $z_0$ is often the RGB image itself. In this paper, generation instead operates in a latent space, where $z_0$ is a multi-view feature tensor later decoded into RGB and depth. Our backbone instantiates this generator with flow matching, a continuous-time objective~\cite{lipman2023flow,ma2024sit} that replaces the discrete denoising chain with a path between the clean latent $z_0$ and a pure noise sample $z_1 \sim \mathcal{N}(0,\mathbf{I})$, where $\mathbf{I}$ is the identity matrix over the flattened latent dimension. For the linear path used in GLD~\cite{jang2026gld}
\begin{equation}\label{eq:linear-path}
z_t = (1-t)z_0 + t z_1,\qquad t \in [0,1].
\end{equation}
The ideal velocity is simply the direction from data to noise
\begin{equation}\label{eq:velocity}
u_t = \frac{\mathrm{d} z_t}{\mathrm{d} t} = z_1 - z_0.
\end{equation}
A network $v_\theta(z_t,t,y)$ with parameters $\theta$ is trained to predict this velocity under the conditional loss
\begin{equation}\label{eq:fm-loss}
\mathcal{L}_{\mathrm{FM}} =
\mathbb{E}_{z_0,\,z_1,\,t}\left[\left\|v_\theta(z_t,t,y) - u_t\right\|_2^2\right].
\end{equation}
At inference, sampling solves the ordinary differential equation $\mathrm{d} z_t / \mathrm{d} t = v_\theta(z_t,t,y)$ backward from the noise $z_1$ at $t{=}1$ to the clean latent $z_0$ at $t{=}0$, typically with an Euler solver.

In our work, the conditioning variable $y$ contains the source images, camera geometry, and projected LiDAR packets. The model therefore does not hallucinate a scene from scratch: it learns a conditional vector field that moves an initially random target-view latent toward the latent consistent with the robot's observations, and the conditioning variable guides that field.

\subsection{2D and 3D Foundation Feature Backbones}\label{subsec:utonia-point-cloud-features}

M3GD composes a dense image-latent backbone with a sparse point-cloud backbone. 

On the 2D side, DA3~\cite{lin2025depthanything3} is a visual-geometry foundation model predicting spatially consistent geometry from one or more visual inputs, with or without known camera poses. M3GD uses not its final depth prediction but the encoded dense image-token representation learned before decoding. Specifically, the encoder maps each image to a spatial feature grid whose cells stay aligned with local image regions, and its depth, ray, and multi-view objectives make those cells geometry-aware as well as appearance-aware. 

These properties motivate our use of DA3 as the image-side latent space for integrating projected point-cloud features, a choice we validate against alternative image backbones in Section~\ref{subsec:findings}.

On the 3D side, Utonia~\cite{zhang2026utonia} is a self-supervised point transformer~\cite{wu2024ptv3,wu2025sonata} trained as a single encoder across heterogeneous point-cloud domains including indoor and outdoor scenes, with no image-feature distillation or other image supervision. This breadth matters for robotic perception because raw point clouds differ sharply in scale, density, sampling pattern, coordinate convention, and optional auxiliary channels such as color or normals. Utonia learns geometry-first per-point descriptors that remain useful across domains rather than specializing to one sensor family. Given a LiDAR sweep $P$, the frozen encoder maps each 3D point to a learned feature attached to its metric location, written $\mathrm{E}_{\mathrm{Utonia}}(P)$.
 
\section{Methodology}\label{sec:method}
\subsection{Overview}\label{subsec:method-overview}
M3GD composes 2D and 3D representations through a projected RGB--LiDAR conditioning pathway (Fig.~\ref{fig:projected-representation}). The frozen DA3 encoder maps the RGB image $I$ to an $H{\times}W$ latent grid $z$, the space in which the inherited GLD backbone generates novel views. In parallel, the frozen Utonia encoder assigns a descriptor to every point of the LiDAR sweep $P$, and the camera projection $\pi$ pools these descriptors into the cells of a second grid $h$ with the same layout. Every cell $q$ thus pairs an image latent $z_q$ with a geometry descriptor $h_q$, while cells without LiDAR support (hatched) carry $h_q=\mathbf{0}$; camera projection alone supplies this correspondence, with neither backbone fine-tuned. Each view-aligned LiDAR \emph{packet} extends $h$ with explicit geometry statistics from the same projected points, and a lightweight residual adapter mixes it into the DA3 grid cell by cell, exposing spatially indexed 3D evidence to the generator while keeping the GLD image latent space, decoders, camera conditioning, and training objective unchanged (Fig.~\ref{fig:m3gd-pipeline}). Section~\ref{subsec:alignment-study} later evaluates whether the frozen image features and projected point-cloud features share measurable spatial structure on this common support. We first specify the inherited GLD backbone and M3GD's operating point (Section~\ref{subsec:geometric-latent-diffusion-gld}), then present the three parts of our method: a projected RGB--LiDAR representation (Section~\ref{subsec:projected-rgb-lidar-representation}), residual LiDAR conditioning (Section~\ref{subsec:residual-conditioning-interface}), and flow-matching training with a configurable inference-step budget (Section~\ref{subsec:training-objectives-and-operating-modes}). Concrete dimensions, channel definitions, and training hyperparameters are isolated in Section~\ref{subsec:implementation-details}.

\subsection{Image-Latent Generative Backbone}\label{subsec:geometric-latent-diffusion-gld}

M3GD builds on GLD~\cite{jang2026gld}, an image-input-only NVS framework that generates DA3 image latents rather than RGB images directly. We specify the inherited latent representation and conditioning mechanism because M3GD retains GLD's generative pipeline and modifies its per-view conditioning feature in Section~\ref{subsec:residual-conditioning-interface}.

\textbf{Latent representation and readout.}
GLD first fixes the space in which generation occurs using one frozen encoder and two frozen decoders
\begin{equation}\label{eq:gld-stage1}
\begin{aligned}
z &= \mathrm{E}_{\mathrm{DA3}}(I) \in \mathbb{R}^{C\times H\times W},\\
\hat I &= \mathrm{D}_{\mathrm{RGB}}(z),
\qquad
\hat D = \mathrm{DPT}(z),
\end{aligned}
\end{equation}
where $z$ is a latent grid of $H{\times}W$ cells aligned with the image patch grid and carrying $C$ channels respectively, $\mathrm{D}_{\mathrm{RGB}}$ is a frozen MAE decoder, $\mathrm{DPT}$ is a frozen depth decoder~\cite{ranftl2021dpt}, and $\hat\cdot$ denotes a model prediction throughout.

The DA3 encoder in eq.~(\ref{eq:gld-stage1}) exposes a hierarchy of four feature levels, indexed $0$--$3$ from shallowest to deepest. Because the frozen encoder can deterministically propagate a shallower feature map to deeper levels, generation needs only to cover the hierarchy up to a chosen boundary level. GLD reports its best quality by explicitly synthesizing levels $0$ and $1$ in a cascade: a level-$1$ model first generates the deeper latent, and a level-$0$ model then generates the shallower latent conditioned on the level-$1$ prediction. M3GD instead adopts a single-model operating point suited to onboard robot deployment and synthesizes only level $0$. This avoids a second generative stage and places LiDAR conditioning at the latent level directly synthesized by the model; the representation analysis in Section~\ref{subsec:alignment-study} also identifies level $0$ as the strongest overall match to projected Utonia features. Section~\ref{subsec:deployment-recipe-ablations} quantifies the deployment cost of synthesizing the deeper level.

\textbf{Multi-view generative conditioning.}
GLD trains a multi-view diffusion transformer~\cite{peebles2023dit,zheng2025rae} to synthesize the joint latent of all $V$ views (Section~\ref{subsec:problem-setup}) at the selected level. The clean latent of the flow-matching formulation is the stack of image encodings, $z_0=\{\mathrm{E}_{\mathrm{DA3}}(I_v)\}_{v=1}^{V}$, and $z_1$ provides the initial noise. Synthesis is conditional on the observed image evidence, camera geometry, and the source or target role of each view. Image evidence is carried by a per-view \emph{conditioning feature} $c_v\in\mathbb{R}^{C\times H\times W}$ defined on the same grid as $z$, while $r_v\in\{0,1\}$ marks the view role as source or target view. The conditioning variable of eq.~(\ref{eq:fm-loss}) is therefore
\begin{equation}\label{eq:gld-y}
\begin{aligned}
y &= \big(\{c_v\}_{v=1}^{V},\ \{(\mathsf{K}_v,T_v)\}_{v=1}^{V},\ \{r_v\}_{v=1}^{V}\big),\\[4pt]
(c_v,\,r_v) &=
\begin{cases}
\big(\mathrm{E}_{\mathrm{DA3}}(I_v),\ 0\big), & v\in\mathcal{V}^\mathrm{src}\\[2pt]
\big(\mathbf{0},\ 1\big), & v\in\mathcal{V}^\mathrm{tgt}
\end{cases}.
\end{aligned}
\end{equation}

The conditioning feature is concatenated with $z_{t,v}$ along the channel axis, giving the per-view network input $[\,c_v\,;\,z_{t,v}\,]\in\mathbb{R}^{2C\times H\times W}$. For $v\in\mathcal{V}^\mathrm{tgt}$, this input reduces to $[\,\mathbf{0}\,;\,z_{t,v}\,]$, where $\mathbf{0}$ is the all-zero tensor with the same shape as $c_v$. Camera geometry $(\mathsf{K}_v,T_v)$ is not concatenated with this input. Instead, each view's Pl\"ucker ray embeddings and broadcast role scalar $r_v$ are patch-embedded and added to the token sequence, while Projective Positional Encoding (PRoPE)~\cite{li2025prope} uses the same intrinsics and extrinsics within the multi-view 3D self-attention~\cite{gao2024cat3d} to couple the $V$ independently noised latents as views of the same scene. The network output $v_\theta(z_t,t,y)$ is a velocity with the same shape as $z_0$. M3GD builds on this multi-view generative backbone and extends its single-modality conditioning to a multimodal one: the per-view conditioning feature $c_v$ is extended with spatially aligned LiDAR evidence on every view, so that target views receive metric 3D evidence, and the multi-view attention couples both sensing modalities across all $V$ views (Section~\ref{subsec:residual-conditioning-interface}).

\subsection{Projected RGB--LiDAR Representation}\label{subsec:projected-rgb-lidar-representation}

The projected RGB--LiDAR representation provides the connection between the LiDAR observation and the image-latent NVS backbone. For each source or target view, a calibrated LiDAR sweep follows two parallel complementary paths: an explicit path that summarizes the metric measurements supporting each image-latent cell, and a learned path that pools frozen Utonia~\cite{zhang2026utonia} per-point descriptors into the same cells.

\textbf{Sweep selection and projection.}
For a view with image timestamp $\tau$, we select the LiDAR sweep $P=\{p_k\}$ whose timestamp $\tau_P$ is nearest $\tau$, which minimizes camera--LiDAR mismatch, and record the signed offset $\Delta\tau = \tau_P - \tau$ in seconds. Each point $p_k$ is transformed into the camera frame through the calibrated LiDAR-to-camera chain (composed via the odometry frame) and projected with the camera intrinsics $\mathsf{K}$, yielding image-plane coordinates $\pi_k$ and camera-frame depth $d_k$. The support set of image-latent cell $q$ is
\begin{equation}\label{eq:support-set}
\mathcal{S}_q=\left\{\,k \,:\, \pi_k \in \text{cell } q,\; d_k>0\,\right\},
\qquad n_q=|\mathcal{S}_q|.
\end{equation}
Projection performs no visibility reasoning: every in-frustum point with positive depth contributes to its cell, including points from mutually occluding surfaces (Section~\ref{sec:limitations}).

\textbf{Explicit geometry statistics.}
With the mean cell depth $\bar d_q=\tfrac{1}{n_q}\sum_{k\in\mathcal{S}_q} d_k$, the explicit component $g_q\in\mathbb{R}^{6}$ of an occupied cell ($n_q{>}0$) is
\begin{equation}\label{eq:packet-explicit}
g_q=\big[\,m_q,\ \log(1{+}\bar d_q),\ \bar d_q^{-1},\ \log(1{+}n_q),\ \textstyle\mathrm{Var}_{k\in\mathcal{S}_q}(d_k),\ \Delta\tau\,\big],
\end{equation}
where $m_q$ is the validity indicator, equal to $1$ in occupied cells, and the variance is taken over the raw depths $d_k$. The offset $\Delta\tau$ is a per-view scalar broadcast into every occupied cell. Empty cells ($n_q{=}0$) carry $g_q=\mathbf{0}$. 

\textbf{Learned point descriptors and packet.}
The learned component averages the frozen Utonia descriptors of the supporting points, and the packet concatenates both components
\begin{equation}\label{eq:packet-learned}
h_q = \frac{1}{n_q}\sum_{k\in\mathcal{S}_q} \mathrm{E}_{\mathrm{Utonia}}(P)_k,
\qquad
\ell_q=[\,g_q;h_q\,],
\end{equation}
where $\mathrm{E}_{\mathrm{Utonia}}(P)_k\in\mathbb{R}^{C_u}$ is the descriptor of point $k$, so $h_q\in\mathbb{R}^{C_u}$. Stacking $\ell_q$ over the $H{\times}W$ latent grid yields the per-view packet $\ell\in\mathbb{R}^{(6+C_u)\times H\times W}$. Cells with $n_q{=}0$ are zero in every channel of both components.

The two components provide complementary views of the same measurement. The explicit component $g_q$ preserves metric range together with the sparsity, density, local depth variation, and timing of the projected sensor support. The learned component $h_q$ supplies a pretrained descriptor of the 3D points contributing to that cell. Camera projection supplies their shared spatial index, while the residual adapter learns how to mix their channels with the image latent.

\textbf{Target-view LiDAR as a geometric query.}
A target packet is computed by exactly the procedure above: sweep selection, projection, and eqs.~(\ref{eq:support-set})--(\ref{eq:packet-learned}) do not depend on whether the view is a source or a target. What differs is how the packet enters the model. A source packet is mixed with the view's image latent, whereas a target view has no image latent, so its packet is the only observation-dependent conditioning of that view (Section~\ref{subsec:residual-conditioning-interface}). 
The construction therefore assumes only that the platform's LiDAR stream covers the queried timestamp,
On the robot (Section~\ref{subsec:real-robot-deployment}), it is the sweep nearest the query timestamp on the logged stream. The query is appearance-free by construction, since the explicit channels are computed from projected point locations alone and Utonia receives only 3D coordinates. Nor does it hand over the depth output: the packet lives on the coarse $H{\times}W$ grid and is empty wherever no point projects, whereas $D^\mathrm{tgt}_j$ is required at every pixel (Section~\ref{subsec:implementation-details}).

\subsection{Residual LiDAR Conditioning}\label{subsec:residual-conditioning-interface}

For every view $v$, GLD receives the channel-wise concatenation of the conditioning feature and the view's noisy latent. Camera poses and intrinsics enter separately through GLD's native camera conditioning (Pl\"ucker ray embeddings and PRoPE-encoded multi-view attention, Section~\ref{subsec:geometric-latent-diffusion-gld}), which M3GD leaves unchanged. In image-only GLD, the conditioning feature is the $c_v$ of eq.~(\ref{eq:gld-y}): the encoded image $\mathrm{E}_{\mathrm{DA3}}(I_v)$ on a source view and the all-zero tensor on a target view, so a target view carries no observation-dependent conditioning beyond its camera geometry. M3GD replaces $c_v$ by the residual update:
\begin{equation}\label{eq:adapter}
\tilde c_v = c_v + \phi\big([\,c_v \,;\, \ell_v \,;\, r_v\,]\big),
\end{equation}
where $\ell_v$ is the view's LiDAR packet, $r_v$ is the role scalar of eq.~(\ref{eq:gld-y}) broadcast over the grid as a single channel, and $\phi:\mathbb{R}^{(C+6+C_u+1)\times H\times W}\rightarrow\mathbb{R}^{C\times H\times W}$ is a zero-initialized residual adapter, so $\tilde c_v$ retains the shape of $c_v$ and the per-view transformer input $[\,\tilde c_v\,;\,z_{t,v}\,]\in\mathbb{R}^{2C\times H\times W}$ is unchanged in shape from image-only GLD. For $v\in\mathcal{V}^\mathrm{tgt}$, eq.~(\ref{eq:adapter}) reduces to $\tilde c_v=\phi([\,\mathbf{0};\ell_v;1\,])$, so the packet is the only observation-dependent conditioning aside from camera geometry. M3GD therefore substitutes $\tilde c_v$ for $c_v$ in eq.~(\ref{eq:gld-y}), giving $y=\big(\{\tilde c_v\}_{v=1}^{V},\ \{(\mathsf{K}_v,T_v)\}_{v=1}^{V},\ \{r_v\}_{v=1}^{V}\big)$ and leaving the other two components untouched. Because $\tilde c_v$ depends on neither the noisy latent nor the flow time $t$, the adapter is evaluated once per synthesis query and reused across all integration steps.

At initialization, the model is exactly the pretrained image-only GLD, and the optimizer opens the LiDAR pathway only as far as it reduces the synthesis loss, a conservative design in the spirit of zero-initialized control branches~\cite{zhang2023controlnet}. The role map let a single shared adapter behave differently on source views, where the packet accompanies a real image latent, and target views, where it does not.

The adapter is intentionally local and lightweight, so it cannot by itself solve cross-view reasoning. Whatever use is made of the packets' spatial structure must happen inside the existing multi-view attention. This keeps the conditioning pathway minimal and makes the ablations of Section~\ref{sec:experiments} interpretable as statements about what the \emph{generative model} extracts from the packets, not about adapter capacity. The same consideration motivates the \emph{Zero} control of Section~\ref{subsubsec:lidar-input-mode}, which keeps the adapter but removes LiDAR content.

\subsection{Flow-Matching Training and Variable-Step Inference}\label{subsec:training-objectives-and-operating-modes}

\textbf{Objective.} M3GD retains GLD's linear-path flow-matching objective with velocity prediction (Section~\ref{subsec:diffusion-models}). The projected LiDAR packets modify only the conditioning variable $y$, so M3GD does not introduce a second prediction objective. The clean latent $z_0\in\mathbb{R}^{V\times C\times H\times W}$ is the stack of the eq.~(\ref{eq:gld-stage1}) encodings over all $V$ views at level $0$, including the target views whose images supply the training supervision but never the conditioning input, and during training we sample intermediate noisy latents along the linear path of eq.~(\ref{eq:linear-path}). Because the transformer denoises all $V$ view latents jointly, the loss of eq.~(\ref{eq:fm-loss}) by default also supervises the source views, whose clean latents the network already receives as conditioning. Our default recipe instead scores only the target-view entries of the joint latent
\begin{equation}\label{eq:fm-tgt}
\mathcal{L}^{\mathrm{tgt}}_{\mathrm{FM}} =
\mathbb{E}_{z_0,z_1,t}\Big[\big\| \Omega^{\mathrm{tgt}}\odot\big(v_\theta(z_t,t,y)-(z_1-z_0)\big)\big\|_2^2\Big],
\end{equation}
where $\Omega^{\mathrm{tgt}}$ is a binary mask selecting the entries of the joint latent with $v\in\mathcal{V}^\mathrm{tgt}$. Section~\ref{subsubsec:lidar-training-mode} compares this target-only loss with the all-view variant and finds that the target-only one performs slightly better.

\textbf{Conditioning dropout and guidance.} During training, the camera conditioning and the LiDAR packets are independently replaced by their null values with probability $0.1$ each, following Classifier-Free Guidance (CFG). The LiDAR null is the all-zero packet, the same value an empty cell already takes in eq.~(\ref{eq:packet-learned}), so a nulled view looks like a view in which no point projected anywhere. Because the model sees such packets on a fraction of training views, it learns to synthesize without LiDAR as well, and a sweep can simply be omitted at deployment when none is available. The \emph{Zero} control of Section~\ref{subsubsec:lidar-input-mode} trains a separate model on this all-zero input for every view and thereby measures the no-LiDAR floor. At inference we apply classifier-free guidance with a jointly nulled condition
\begin{equation}\label{eq:cfg}
v^{\mathrm{cfg}}_\theta(z_t,t) = v_\theta(z_t,t,\varnothing) + s\big(v_\theta(z_t,t,y)-v_\theta(z_t,t,\varnothing)\big),
\end{equation}
where $\varnothing$ nulls the camera conditioning and all LiDAR packets simultaneously while retaining the source image latents, and $s{=}1.5$.

\textbf{Inference.} All $V$ view latents are initialized from noise $z_1\sim\mathcal{N}(0,\mathbf{I})$ and integrated jointly: a $K$-step Euler solver applies the guided velocity of eq.~(\ref{eq:cfg}) from $t{=}1$ to $t{=}0$ with step size $1/K$. The source-view latents are generated alongside the targets and discarded. Each target's generated level-$0$ latent is then propagated through the remaining frozen DA3 encoder blocks to the deeper feature levels, from which the frozen decoders recover RGB and depth. All choices of $K$ use the same model and weights: reducing $K$ lowers the number of network evaluations, while increasing it more finely resolves the learned flow trajectory. In particular, $K{=}1$ applies one Euler update to the same learned velocity field. Section~\ref{subsec:deployment-runtime} measures this quality--cost trade-off on the robot platform.
 
\section{Experimental Setup}\label{sec:expsetup}
This section fixes the protocol shared by all experiments in Section~\ref{sec:experiments}: implementation (Section~\ref{subsec:implementation-details}), dataset and metrics (Section~\ref{subsec:dataset-metrics}), baselines (Section~\ref{subsec:baselines}), and the real-robot platform (Section~\ref{subsec:robot-platform}).

\subsection{Implementation Details}\label{subsec:implementation-details}
\textbf{Backbones.}
We leverage the released GLD checkpoint~\cite{jang2026gld}, whose frozen encoder and decoders are described in Section~\ref{subsec:geometric-latent-diffusion-gld}. Each DA3 feature level synthesized by the diffusion model is a $1536$-channel map ($C{=}1536$) on the image patch grid. LiDAR points are encoded by the released Utonia model~\cite{zhang2026utonia}, kept frozen and maps point coordinates to $1224$-channel per-point features ($C_u{=}1224$).

\textbf{LiDAR-to-grid conditioning.}
All cameras follow the OpenCV convention. Each sweep is encoded by Utonia once, and its per-point features are shared by all simultaneous views that select it; projection and pooling remain per-view. Points are projected with intrinsics matching the dataloader's crop-and-resize and pooled onto the DA3 latent grid, of size $H{\times}W{=}16{\times}24$ cells for the $224{\times}336$ evaluation images, so a fully supported packet constrains $384$ cells against the $75{,}264$ pixels at which depth is predicted, and cells without projected points are empty. The explicit component $g$ contains the six channels of eq.~(\ref{eq:packet-explicit}) and the learned component $h$ contains the $1224$ pooled Utonia channels, so the packet $\ell=[g;h]$ is a $1230{\times}H{\times}W$ conditioning tensor.

\textbf{Residual adapter.} We inject LiDAR conditioning at DA3-Base level-$0$, which is both the level synthesized by the image diffusion backbone and the strongest overall match in the representation analysis of Section~\ref{subsec:findings}. LiDAR conditioning therefore enters at the same level at which generation occurs; the deeper feature levels are reconstructed only after sampling, when the frozen encoder propagates the generated level-$0$ latent (Section~\ref{subsec:training-objectives-and-operating-modes}). The adapter $\phi$ is a two-layer $1{\times}1$ convolution with SiLU activation and a zero-initialized final layer, with channels $1230+1536+1 \rightarrow 256 \rightarrow 1536$. The extra channel is the source/target role map. It adds $\sim1.1$M parameters, roughly $0.2\%$ of the transformer.

\textbf{Training and inference.} We fine-tune from the released GLD checkpoint with AdamW for $50$k optimization steps at global batch size $32$, constant learning rate $5{\times}10^{-5}$, bf16 precision, EMA decay $0.9995$, and the independent camera and LiDAR conditioning dropout of Section~\ref{subsec:training-objectives-and-operating-modes}. Each clip provides $V{=}4$ shuffled views for both training and evaluation, with $1$--$3$ source views per example during training and $2$ at evaluation, and frame gaps of $1$--$20$ frames at the native camera rate. Inference follows the guided Euler procedure of Section~\ref{subsec:training-objectives-and-operating-modes} with a $K{=}50$-step flow-matching schedule; the deployment study varies $K$ without retraining. All unlisted settings follow the GLD defaults.

\subsection{Dataset and Evaluation Metrics}\label{subsec:dataset-metrics}
\textbf{Dataset.} We use GrandTour~\cite{frey2026grandtour}, an ANYmal-D quadruped carrying the Boxi sensor payload that traverses indoor facilities, urban outdoor scenes, and natural terrain. GrandTour suits M3GD on three counts. Its Boxi payload provides the hardware-synchronized, calibrated camera--LiDAR streams and reliable poses (RTK-GNSS and total-station tracking) that packet construction requires; as the largest open-access legged-robotics dataset, it offers the scale for fine-tuning and the environment diversity our evaluation draws from; and its ego-centric 6-DoF viewpoints match our deployment setting (Section~\ref{subsec:robot-platform}), whereas driving datasets such as nuScenes~\cite{caesar2020nuscenes} pair camera and LiDAR only under near-planar road motion and internet-scale NVS collections provide no LiDAR at all. We use the three HDR cameras (front, left, right) and the motion-undistorted Hesai LiDAR. We train and evaluate in temporal mode, where source and target views are sampled from a single camera stream. Packet construction and view-sampling details follow Section~\ref{subsec:implementation-details}.

As GrandTour provides no official train/evaluation split, we partition its $48$ recordings into $40$ for training and $8$ for evaluation, \emph{with no recording overlap between the two splits}. Training uses all three HDR cameras (front, left, right) and the LiDAR; evaluation uses the front camera and LiDAR only, with two conditioning and two target views per clip and $300$ clips sampled per recording ($2{,}400$ clips in total). Each target view may use its co-located LiDAR observation as the LiDAR query defined in Section~\ref{subsec:projected-rgb-lidar-representation}. The target RGB image is never provided to the model. The evaluation recordings span eight representative scenarios, namely ice cave, alpine snow, urban plaza, forest, industrial bunker, warehouse/garage, railway, and construction, yielding a diverse benchmark that stresses both texture-rich and geometry-dominant settings.

\textbf{Data for the representation study.} The representation-compatibility study of Section~\ref{subsec:alignment-study} involves no training, so it is not bound to the split above: it uses all $48$ recordings and all three HDR camera streams, $599{,}922$ images in total. Of these, $599{,}774$ are processed; the remaining $148$ contain fewer than three valid LiDAR-supported cells and are excluded.

\textbf{Evaluation metrics.} We assess synthesized views along three axes: 2D appearance, 2D geometry (depth), and 3D reconstruction. All metrics are computed on \emph{target} views only, since source views are model inputs, and are averaged over the target views of the $2{,}400$ evaluation clips at a fixed seed. Where noted, we additionally report a \emph{hard subset} of $n{=}1{,}407$ clips whose maximum camera baseline between views exceeds $1$\,m. For appearance, we report PSNR, SSIM~\cite{wang2004ssim}, and LPIPS~\cite{zhang2018lpips} on the rendered RGB. For depth, GrandTour provides no dense ground-truth depth, and several baselines emit no depth at all, so we evaluate all methods through a single frozen estimator: each rendered target view $\hat I^\mathrm{tgt}_j$ is re-encoded by the frozen DA3 encoder $\mathrm{E}_\mathrm{DA3}$ and decoded by the DPT head (Section~\ref{subsec:implementation-details}), and we compare $\mathrm{DPT}(\mathrm{E}_\mathrm{DA3}(\hat I^\mathrm{tgt}_j))$ against the pseudo-ground truth $\mathrm{DPT}(\mathrm{E}_\mathrm{DA3}(I^\mathrm{tgt}_j))$ using absolute relative error (AbsRel), RMSE, and the inlier ratio $\delta{<}1.25$~\cite{eigen2014depth}. Because depth is derived uniformly from rendered RGB rather than read from any method's native output, these numbers quantify appearance--geometry self-consistency rather than absolute metric depth. For 3D reconstruction, we lift each target view to 3D using this per-view depth and its camera, and measure cross-view consistency via MEt3R~\cite{asim2025met3r} and 3D reprojection error~\cite{du2026videogpa}.

\subsection{Baselines}\label{subsec:baselines}
We compare against representative NVS methods from two paradigms, all operating from sparse-view input to match our task setting
\begin{itemize}
    \item \emph{Diffusion-based approaches:} Matrix3D~\cite{lu2025matrix3d}, MVGenMaster~\cite{cao2025mvgenmaster}, and GLD~\cite{jang2026gld}, the latter both off-the-shelf and fine-tuned on the GrandTour training set at DA3 level~$0$. Throughout the tables, $^{*}$ marks a model fine-tuned on the GrandTour training set and $^{\dagger}$ trained on the DA3 level-$0$ feature, so the fine-tuned baseline is written GLD$^{*\dagger}$
    \item \emph{Feed-forward splat-based approaches:} MVSplat~\cite{chen2024mvsplat}, DepthSplat~\cite{xu2025depthsplat}, and PixelSplat~\cite{charatan2024pixelsplat}.
\end{itemize}

\subsection{Real-Robot Platform}\label{subsec:robot-platform}
\begin{figure}[t]
    \centering
    \includegraphics[width=.95\columnwidth]{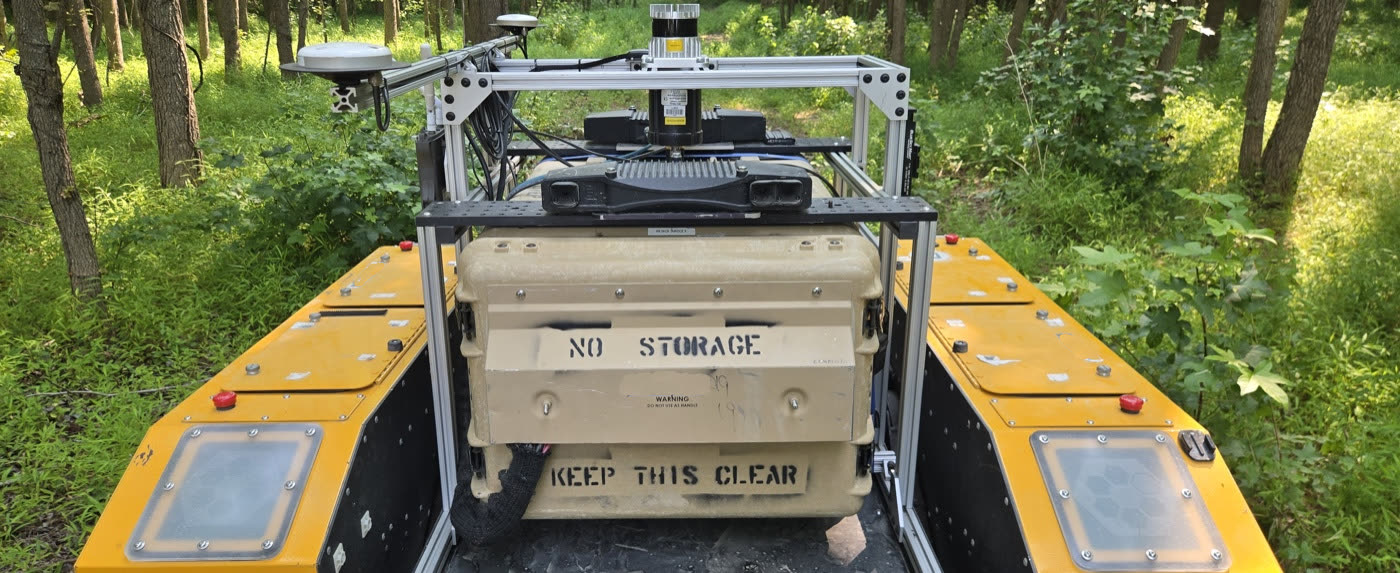}
    \vspace{-10pt}
    \caption{Clearpath Warthog platform used for the real-robot deployment.}
    \label{fig:warthog-platform}
    \vspace{-10pt}
\end{figure}
For deployment and runtime experiments, we use a wheeled ground robot, a Clearpath Warthog (Fig.~\ref{fig:warthog-platform}), carrying a calibrated MultiSense S27 RGB-D sensor and a $360$-degree Ouster LiDAR. The onboard computer is a Neousys RGS-8805GC with an NVIDIA RTX A6000 GPU. Runtime is measured in a persistent loaded process over four timed queries after one warm-up query. Because the model jointly decodes all $N+M$ views, latency and memory cover the complete call; latency per target amortizes that call over the $M$ held-out targets. CUDA synchronization brackets source encoding, flow generation, and output decoding. We use peak PyTorch reserved memory as the primary GPU-memory metric because it is the memory held by PyTorch's CUDA caching allocator and therefore unavailable to other processes.

\section{Results}\label{sec:experiments}
Under the protocol of Section~\ref{sec:expsetup}, we first test the representation-level premise of M3GD (Section~\ref{subsec:alignment-study}). We then present the baseline comparison and mechanism ablations (Sections~\ref{subsec:baseline} and~\ref{subsec:ablation}), qualitative analysis (Section~\ref{subsec:qual}), and the deployment-recipe, real-robot, and runtime results (Sections~\ref{subsec:deployment-recipe-ablations}--\ref{subsec:deployment-runtime}).

\subsection{Projected 2D--3D Representation Compatibility}\label{subsec:alignment-study}
Having established M3GD's projected LiDAR conditioning pathway, we test its representation-level premise: whether independently trained image and point-cloud foundation encoders organize corresponding scene regions similarly after camera projection. Projection places both representations on the same image-latent grid, so corresponding cells describe the same scene region; however, the feature vectors at those cells live in independently learned spaces of different dimensions and need not be similar in any coordinate-wise sense. Utonia is pretrained purely on point clouds, with no image-feature distillation or other image supervision (Section~\ref{subsec:utonia-point-cloud-features}), so any measured compatibility with DA3 emerges without explicit image--point alignment training.

All quantities in this study are computed per image; we omit the image index. Let $\mathcal{Q}=\{q:n_{q}>0\}$ denote the image-grid cells with valid projected LiDAR support under the support sets of eq.~(\ref{eq:support-set}), $z_{q}^{(l)}$ the image-encoder feature at cell $q$, and $h_{q}$ the pooled point descriptor of eq.~(\ref{eq:packet-learned}). On these shared cells, we form the image-feature field at DA3 level $l$ and the projected Utonia field
\begin{equation}\label{eq:alignment-fields}
\begin{aligned}
Z^{(l)} &= [\,z_{q}^{(l)}\,]_{q\in\mathcal{Q}}
    \in \mathbb{R}^{|\mathcal{Q}|\times C_l},\\
U &= [\,h_{q}\,]_{q\in\mathcal{Q}}
    \in \mathbb{R}^{|\mathcal{Q}|\times C_u},
\end{aligned}
\end{equation}
The question is one of compatibility between two \emph{learned} feature spaces, the image features $z_q^{(l)}$ and the point-cloud features $h_q$, so of the packet $\ell_q=[\,g_q;h_q\,]$ of eq.~(\ref{eq:packet-learned}) only the learned component enters the study: the explicit component $g_q$ is a handcrafted vector of depth statistics rather than a learned representation, so there is no feature space on its side to compare. Since $C_l \neq C_u$ in general, we measure agreement in the spatial relations encoded across shared cells rather than comparing features entry-wise. We summarize this structural compatibility with
\begin{equation}\label{eq:alignment-diagnostics}
\mathcal{A}(Z,U)=
\big[\rho_{\mathrm{PCA}},\ \mathrm{CKA},\ \mathrm{SVCCA}_{20},\
\ \mathrm{Jaccard}_{10},\ \rho_{\mathrm{aff}}\big].
\end{equation}

\subsubsection{GrandTour Protocol and Diagnostics}\label{subsec:protocol}

\begin{table*}[!t]
\centering
\footnotesize
\setlength{\tabcolsep}{6pt}
\caption{
Feature-structure similarity on GrandTour.
All metrics are computed on shared $16\times24$ image-grid cells with valid projected LiDAR support.
The study covers $599{,}774$ successfully processed images from $48$ recordings and three HDR camera streams.
\textbf{Best} and \underline{second-best} are highlighted within each comparison block.
}
\label{tab:feature_alignment}
\begin{tabular}{lccccc}
\toprule
Feature pair
& PCA ch. corr. $\uparrow$
& CKA $\uparrow$
& SVCCA-$20$ $\uparrow$
& Jaccard@$10$ $\uparrow$
& Affinity corr. $\uparrow$ \\
\midrule

\multicolumn{6}{l}{\emph{DA3-Base feature level vs.\ Utonia}} \\
DA3-Base level-$0$--Utonia
& \underline{$0.6412$}
& $\mathbf{0.6590}$
& $\mathbf{0.6040}$
& \underline{$0.3551$}
& \underline{$0.5737$} \\
DA3-Base level-$1$--Utonia
& $0.6212$
& \underline{$0.6451$}
& \underline{$0.5911$}
& $0.3402$
& $\mathbf{0.5997}$ \\
DA3-Base level-$2$--Utonia
& $0.6347$
& $0.5459$
& $0.5789$
& $0.3460$
& $0.5372$ \\
DA3-Base level-$3$--Utonia
& $\mathbf{0.6426}$
& $0.4688$
& $0.5796$
& $\mathbf{0.3565}$
& $0.4580$ \\

\midrule
\multicolumn{6}{l}{\emph{Image backbone comparison vs.\ Utonia}} \\
DA3-Base level-$0$--Utonia
& $\mathbf{0.6412}$
& $\mathbf{0.6590}$
& $\mathbf{0.6040}$
& $\mathbf{0.3551}$
& $\mathbf{0.5737}$ \\
DINOv2--Utonia
& $0.6155$
& \underline{$0.6363$}
& $0.5792$
& $0.2690$
& \underline{$0.5659$} \\
VGGT--Utonia
& \underline{$0.6398$}
& $0.4284$
& \underline{$0.5826$}
& \underline{$0.3497$}
& $0.3424$ \\

\midrule
\multicolumn{6}{l}{\emph{Cross-modal alignment vs.\ image-only reference}} \\
DA3-Base level-$0$--Utonia
& \underline{$0.6412$}
& \underline{$0.6590$}
& \underline{$0.6040$}
& $\mathbf{0.3551}$
& \underline{$0.5737$} \\
DA3-Base level-$0$--DINOv2
& $\mathbf{0.6973}$
& $\mathbf{0.7314}$
& $\mathbf{0.6793}$
& \underline{$0.3544$}
& $\mathbf{0.6501}$ \\

\bottomrule
\end{tabular}
\vspace{-10pt}
\end{table*}

\begin{figure}
    \centering
    \scriptsize
    \makebox[0.25\columnwidth][c]{RGB}%
    \makebox[0.25\columnwidth][c]{DA3-Base Level-$0$ PCA}%
    \makebox[0.25\columnwidth][c]{Utonia PCA}%
    \makebox[0.25\columnwidth][c]{DINOv2 PCA}\\[0.2em]
    \includegraphics[width=\columnwidth]{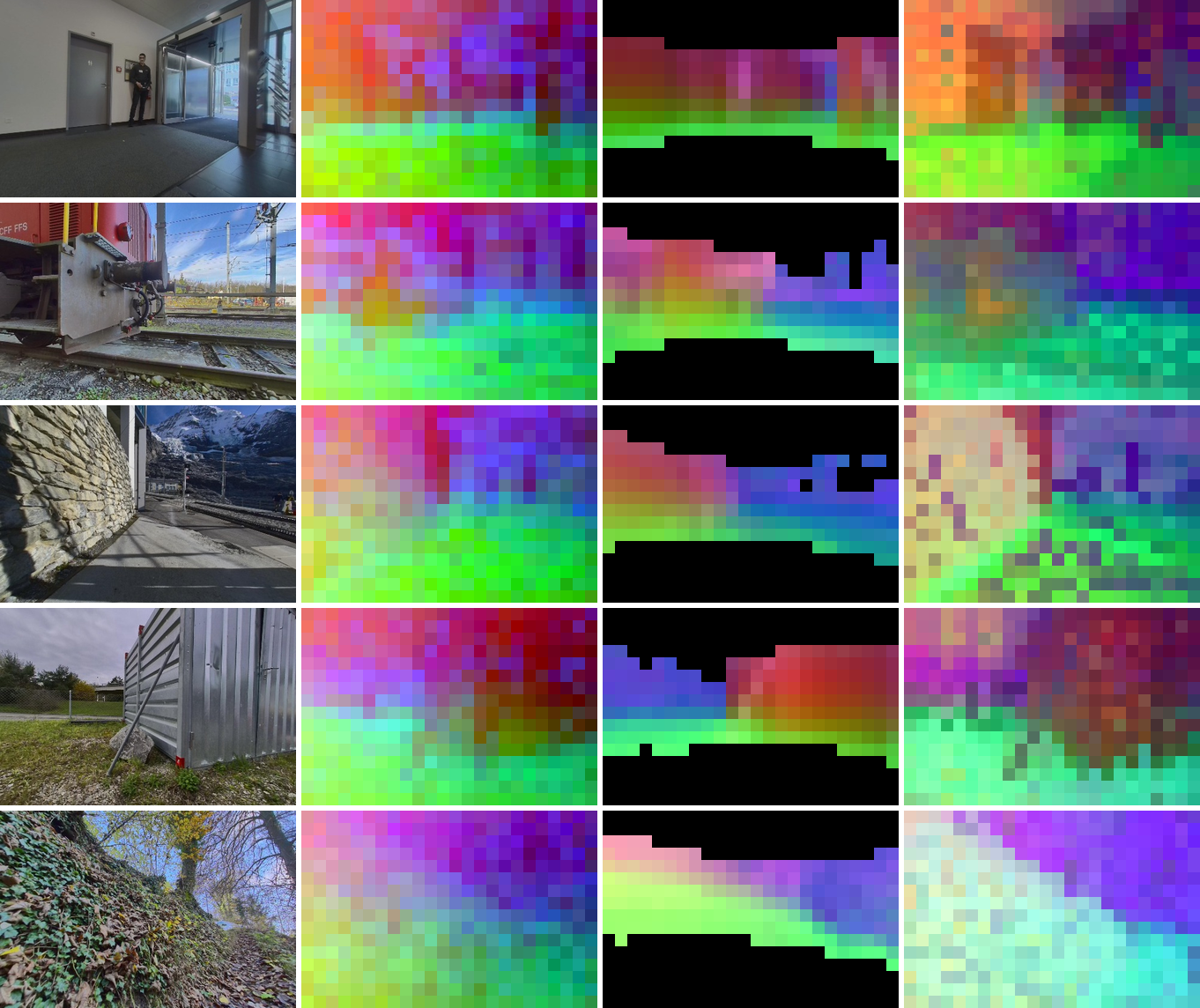}
    \vspace{-10pt}
    \caption{
    Qualitative feature-structure visualization on GrandTour.
    For each camera image, we show the RGB input, the DA3-Base level-$0$ PCA pseudo-color map, projected Utonia features, and DINOv2 final-layer image features.
    For each image, DA3-Base, Utonia, and DINOv2 features are independently projected to three local PCA components on the shared valid grid cells. The Utonia and DINOv2 coordinates are aligned to DA3-Base level-$0$ using orthogonal Procrustes.
    Black regions in the Utonia column indicate cells without valid projected LiDAR support.
    Projected Utonia features recover large-scale geometric partitions similar to DA3-Base level-$0$ on LiDAR-supported cells, while DINOv2 provides an image-only reference.
    }
    \label{fig:feature_alignment_pca}
    \vspace{-10pt}
\end{figure}
We instantiate eqs.~(\ref{eq:alignment-fields})--(\ref{eq:alignment-diagnostics}) on GrandTour~\cite{frey2026grandtour}, chosen for the reasons given in Section~\ref{subsec:dataset-metrics} and whose scale additionally supports population-level alignment statistics, on the image set described there; the $16\times24$ image-latent grid of the $224\times336$ images (Section~\ref{subsec:implementation-details}) provides the common spatial support for cross-modal comparison.

For each image, we compute: (i) DA3-Base~\cite{lin2025depthanything3} encoder features from levels $0$--$3$; (ii) DINOv2~\cite{oquab2024dinov2learningrobustvisual} final-layer patch features as an image-only reference; (iii) VGGT image features as an additional geometry-aware image-backbone reference; and (iv) frozen Utonia~\cite{zhang2026utonia} per-point features from the associated LiDAR sweep. The Utonia features are projected into the camera view and average-pooled onto the same grid. All comparisons are restricted to cells in $\mathcal{Q}$.

We evaluate three comparison groups. First, DA3-Base levels $0$--$3$ are compared against Utonia to measure level-wise compatibility. Second, DA3-Base level-$0$--Utonia is compared against DINOv2--Utonia and VGGT--Utonia to test whether DA3 is more or less compatible with Utonia than alternative image backbones. Third, DA3-Base level-$0$--Utonia is compared with DA3-Base level-$0$--DINOv2, which provides an image-only reference scale rather than a competing conditioning signal.

For $\rho_{\mathrm{PCA}}$, we independently project each representation in each image to three local PCA components, align the pseudo-color coordinates with orthogonal Procrustes, and report their mean channel correlation. This diagnostic summarizes low-dimensional spatial consensus rather than raw-feature equivalence. On the same supported cells before PCA, we compute linear CKA~\cite{kornblith2019cka}, mean SVCCA correlation over up to $20$ available directions~\cite{raghu2017svcca}, Jaccard@$10$ agreement between cosine-nearest-neighbor sets, and the correlation $\rho_{\mathrm{aff}}$ between pairwise cosine-affinity matrices. Table~\ref{tab:feature_alignment} reports macro averages over all successfully processed images; when fewer than ten non-self neighbors are available, Jaccard uses the available set. Representative PCA visualizations appear in Fig.~\ref{fig:feature_alignment_pca}.

\subsubsection{Findings}\label{subsec:findings}
\textbf{DA3-Base level-$0$ is the most compatible feature level.}
Among the DA3-Base hierarchy, level-$0$ shows the strongest overall compatibility with projected Utonia features. It achieves the highest CKA and SVCCA-$20$, while ranking second on PCA channel correlation, Jaccard@$10$, and affinity correlation. Although higher levels are competitive on isolated low-dimensional or neighborhood diagnostics, their raw-feature-space alignment is weaker, especially in CKA. This result is consistent with conditioning M3GD at level $0$.

\textbf{DA3 is the strongest tested image-side match to Utonia.}
Under the same grid and LiDAR-supported cells, DA3-Base level-$0$ obtains the highest CKA, SVCCA-$20$, Jaccard@$10$, and affinity correlation among the image backbones compared against Utonia. The PCA visualizations in Fig.~\ref{fig:feature_alignment_pca} likewise show that projected Utonia and DA3 features often partition co-located scene regions similarly. These results place DA3 ahead of DINOv2 and VGGT for the tested Utonia pairing.

\textbf{Cross-modal similarity is substantial but not image--image equivalence.}
As expected, the image-only DA3-Base level-$0$--DINOv2 reference is stronger than the cross-modal DA3-Base level-$0$--Utonia pair on most raw-feature diagnostics, including CKA, SVCCA-$20$, and affinity correlation. The cross-modal pair nevertheless remains on the same broad scale and slightly exceeds the image-only reference on Jaccard@$10$. We therefore interpret the evidence as shared spatial and geometric structure between the tested independently trained encoders, not coordinate-wise feature equivalence.

\textbf{Implication.}
These findings support the representation-compatibility premise underlying M3GD and are consistent with exposing projected Utonia features to the generator at DA3-Base level-$0$. M3GD does not separately pretrain a cross-modal translator or introduce an auxiliary alignment objective; its residual adapter is learned end-to-end through the NVS objective. Representation compatibility and downstream utility are separate questions: the baseline comparison (Section~\ref{subsec:baseline}) and the mechanism ablations (Section~\ref{subsubsec:lidar-input-mode}) test whether real projected LiDAR packets improve synthesis beyond adapter capacity alone.
 
\subsection{Baseline Comparison}\label{subsec:baseline}
Table~\ref{tab:baseline} compares M3GD with diffusion-based and feed-forward splat-based NVS methods on the GrandTour evaluation split. The fine-tuned GLD row is the closest image-only baseline: it uses the same robot data and image-latent backbone but does not use the projected LiDAR representation.

\begin{table*}
\centering
\caption{Baseline comparison on the GrandTour evaluation split ($2{,}400$ clips). $\uparrow$/$\downarrow$ indicate that higher/lower is better; \textbf{best} and \underline{second-best} per column are highlighted. $^{*}$\,fine-tuned on the GrandTour training set; $^{\dagger}$\,trained on the DA3 level-$0$ feature.
}
\label{tab:baseline}
\footnotesize
\setlength{\tabcolsep}{6pt}
\renewcommand{\arraystretch}{1.15}
\begin{tabular}{l ccc ccc cc}
\toprule
& \multicolumn{3}{c}{2D Appearance} & \multicolumn{3}{c}{2D Geometry (Depth)} & \multicolumn{2}{c}{3D Reconstruction} \\
\cmidrule(lr){2-4} \cmidrule(lr){5-7} \cmidrule(lr){8-9}
Method & PSNR\,$\uparrow$ & SSIM\,$\uparrow$ & LPIPS\,$\downarrow$
       & AbsRel\,$\downarrow$ & RMSE\,$\downarrow$ & $\delta{<}1.25$\,$\uparrow$
       & MEt3R\,$\downarrow$ & Reproj.\,$\downarrow$ \\
\midrule
\multicolumn{9}{l}{\emph{Diffusion-based methods}} \\
Matrix3D~\cite{lu2025matrix3d}            & $16.90$ & $0.374$ & $0.502$ & $0.282$ & $0.570$ & $0.587$ & $\mathbf{0.184}$ & $0.071$ \\
MVGenMaster~\cite{cao2025mvgenmaster}     & $17.45$ & $0.406$ & $0.446$ & $0.285$ & $0.568$ & $0.629$ & $0.243$ & $0.072$ \\
GLD~\cite{jang2026gld}                    & $17.92$ & $0.414$ & $0.450$ & $0.303$ & $0.602$ & $0.592$ & $0.283$ & $0.064$ \\
GLD$^{*\dagger}$~\cite{jang2026gld}       & \underline{$21.62$} & \underline{$0.531$} & \underline{$0.302$} & \underline{$0.173$} & \underline{$0.354$} & \underline{$0.777$} & $0.191$ & $\mathbf{0.059}$ \\
\midrule
\multicolumn{9}{l}{\emph{Splat-based methods}} \\
MVSplat~\cite{chen2024mvsplat}            & $18.57$ & $0.450$ & $0.484$ & $0.318$ & $0.769$ & $0.475$ & $0.212$ & $0.062$ \\
DepthSplat~\cite{xu2025depthsplat}        & $17.98$ & $0.415$ & $0.505$ & $0.355$ & $0.836$ & $0.436$ & $0.234$ & $0.066$ \\
PixelSplat~\cite{charatan2024pixelsplat}  & $18.69$ & $0.449$ & $0.497$ & $0.320$ & $0.768$ & $0.472$ & $0.234$ & \underline{$0.061$} \\
\midrule
\textbf{M3GD}$^{*\dagger}$ \textbf{(Ours)} & $\mathbf{22.33}$ & $\mathbf{0.549}$ & $\mathbf{0.283}$ & $\mathbf{0.145}$ & $\mathbf{0.314}$ & $\mathbf{0.810}$ & \underline{$0.189$} & $\mathbf{0.059}$ \\
\bottomrule
\end{tabular}
\end{table*}
 
\textbf{2D appearance.} Across the eight metrics, M3GD ranks first on seven---with GLD$^{*\dagger}$ matching its best reprojection error at the reported precision---and second on MEt3R. On 2D appearance it improves over every baseline, but the margin over GLD$^{*\dagger}$ is modest. Since the two rows share backbone, training data and DA3 level-$0$ features and differ only in projected-LiDAR conditioning, this isolates the effect of LiDAR on aggregate pixel fidelity, which is incremental. The feed-forward splat baselines trail both diffusion families: the strongest splat PSNR (PixelSplat) is $3.6$\,dB below M3GD.
 
\textbf{2D geometry.} The depth metrics show a larger relative gap: M3GD lowers AbsRel by $16\%$ ($0.145$ vs.\ $0.173$) and RMSE by $11\%$ ($0.314$ vs.\ $0.354$), and raises the $\delta{<}1.25$ inlier ratio from $0.777$ to $0.810$, while all three splat baselines exceed $0.31$ AbsRel and $0.76$ RMSE. Because all numbers are computed uniformly from the rendered RGB via the RGB readout (see \emph{Depth readout comparison} below), the improvement reflects sharper geometry-relevant structure in the synthesized views and is consistent with M3GD's access to target-view LiDAR geometry that image-only baselines cannot exploit.
 
\textbf{3D reconstruction.} M3GD ties GLD$^{*\dagger}$ for the lowest reprojection error at the reported precision ($0.059$) and slightly outperforms the strongest splatting result, PixelSplat at $0.061$. On MEt3R, M3GD ranks second, behind Matrix3D and slightly ahead of GLD$^{*\dagger}$. Because MEt3R measures cross-view consistency rather than image fidelity, Matrix3D's strong MEt3R score despite its low PSNR and SSIM suggests that consistent but visually under-detailed predictions can perform well on this metric. The near tie between M3GD and GLD$^{*\dagger}$ indicates that LiDAR conditioning improves per-view geometry more clearly than cross-view agreement.
 \begin{table}[t]
\centering
\caption{Depth readout comparison, using identical latent predictions as Table~\ref{tab:baseline}. The \emph{native readout} scores depth decoded directly from the synthesized latent (GLD-default); the \emph{RGB readout} first decodes to RGB and re-encodes through the DA3 encoder (protocol of Table~\ref{tab:baseline}). Both are scored against the same pseudo-ground truth. Notation follows Table~\ref{tab:baseline}; \textbf{best} within each readout.}
\label{tab:depth_protocol}
\footnotesize
\setlength{\tabcolsep}{6pt}
\renewcommand{\arraystretch}{1.15}
\begin{tabular}{l ccc}
\toprule
Method & AbsRel\,$\downarrow$ & RMSE\,$\downarrow$ & $\delta{<}1.25$\,$\uparrow$ \\
\midrule
\multicolumn{4}{l}{\emph{Native readout (GLD-default)}} \\
GLD                               & $0.358$ & $0.747$ & $0.478$ \\
GLD$^{*\dagger}$                  & $0.248$ & $0.500$ & $0.565$ \\
\textbf{M3GD}$^{*\dagger}$ \textbf{(Ours)} & $\mathbf{0.227}$ & $\mathbf{0.472}$ & $\mathbf{0.579}$ \\
\midrule
\multicolumn{4}{l}{\emph{RGB readout (protocol of Table~\ref{tab:baseline})}} \\
GLD                               & $0.303$ & $0.602$ & $0.592$ \\
GLD$^{*\dagger}$                  & $0.173$ & $0.354$ & $0.777$ \\
\textbf{M3GD}$^{*\dagger}$ \textbf{(Ours)} & $\mathbf{0.145}$ & $\mathbf{0.314}$ & $\mathbf{0.810}$ \\
\bottomrule
\end{tabular}

\end{table}
\textbf{Depth readout comparison.}
The depth columns of Table~\ref{tab:baseline} depend on a readout choice, which we validate here. GLD's \emph{native readout} feeds the synthesized latent $\hat z_0$, a sample in the frozen DA3 encoder space (Section~\ref{subsec:geometric-latent-diffusion-gld}), to DA3's DPT head ($\mathrm{DPT}(\hat z_0)$). Our \emph{RGB readout} instead decodes to RGB and re-encodes through the frozen encoder ($\mathrm{DPT}(\mathrm{E}_\mathrm{DA3}(\mathrm{D}_\mathrm{RGB}(\hat z_0)))$), the only readout defined for the RGB-only baselines. Table~\ref{tab:depth_protocol} compares the two on identical latent predictions.

The RGB readout reports uniformly lower error for M3GD$^{*\dagger}$. We attribute this to smoothing rather than a more accurate readout: re-encoding projects the prediction onto the estimator's natural-image manifold, suppressing latent synthesis error, whereas the native readout exposes it directly and is therefore the stricter protocol. Crucially, the ranking is unchanged: M3GD$^{*\dagger}$ leads on all three metrics under both readouts, with GLD$^{*\dagger}$ second and GLD last, so the gap isolates the projected-LiDAR conditioning either way. Finally, since several baselines lack a native depth readout, we adopt the RGB readout for all methods to ensure a fair comparison.

\textbf{Inference cost.}
Table~\ref{tab:gld-vs-m3gd} reports the sampling cost of the diffusion methods. M3GD is not only the most accurate method in Table~\ref{tab:baseline} but also the cheapest to run, with the lowest latency and the lowest peak reserved memory of all diffusion methods. The cleanest comparison is against GLD, which shares M3GD's resolution and sampling budget: M3GD is $1.7{\times}$ faster and uses $33\%$ less memory, largely because GLD's native two-stage cascade evaluates the denoiser twice per sampling step whereas M3GD samples in a single pass. Matrix3D and MVGenMaster run in their original configurations at higher resolutions ($512{\times}512$ and $384{\times}576$ against $336{\times}224$), and Matrix3D additionally samples once per target view; even so, M3GD is $7.6{\times}$ faster than Matrix3D, matches MVGenMaster's latency within run-to-run variation, and uses roughly $40\%$ less memory than either. The reported M3GD latency is moreover conservative: it includes projecting the target-view LiDAR and running the frozen Utonia encoder, both of which execute once per clip and amortize over all $M$ target views. In a streaming deployment the packet can be computed as the scan arrives rather than at query time, lowering the per-query cost further.

\begin{table}[t]
\centering
\caption{Sampling cost of the diffusion methods on the GrandTour evaluation split. All methods use $N{=}2$ source views and $M{=}2$ target views. Latency is measured per target pair over the full sampling procedure, including the VAE decode to RGB and, for GLD, both stages of its native cascade; for M3GD it also includes generation of the LiDAR packet. Values are averaged over three runs ($\pm$ standard deviation); peak reserved GPU memory is measured on a single NVIDIA RTX A5000, following the protocol of Table~\ref{tab:deployment_scaling}. \textbf{Best} per column.}
\label{tab:gld-vs-m3gd}
\footnotesize
\setlength{\tabcolsep}{6pt}
\renewcommand{\arraystretch}{1.15}
\begin{tabular}{l cc}
\toprule
Method & Latency (s)\,$\downarrow$ & Memory (GiB)\,$\downarrow$ \\
\midrule
Matrix3D~\cite{lu2025matrix3d}        & $77.7 \pm 0.1$ & $11.0$ \\
MVGenMaster~\cite{cao2025mvgenmaster} & $10.3 \pm 0.1$ & $10.5$ \\
GLD~\cite{jang2026gld}     & $17.3 \pm 0.8$ & $9.7$ \\
\textbf{M3GD}$^{*\dagger}$ \textbf{(Ours)} & $\mathbf{10.2 \pm 0.2}$ & $\mathbf{6.5}$ \\
\bottomrule
\end{tabular}
\vspace{-15pt}
\end{table}

\subsection{Qualitative Analysis}\label{subsec:qual}
\begin{figure*}[p]
    \centering
    \scriptsize
    \includegraphics[width=\textwidth]{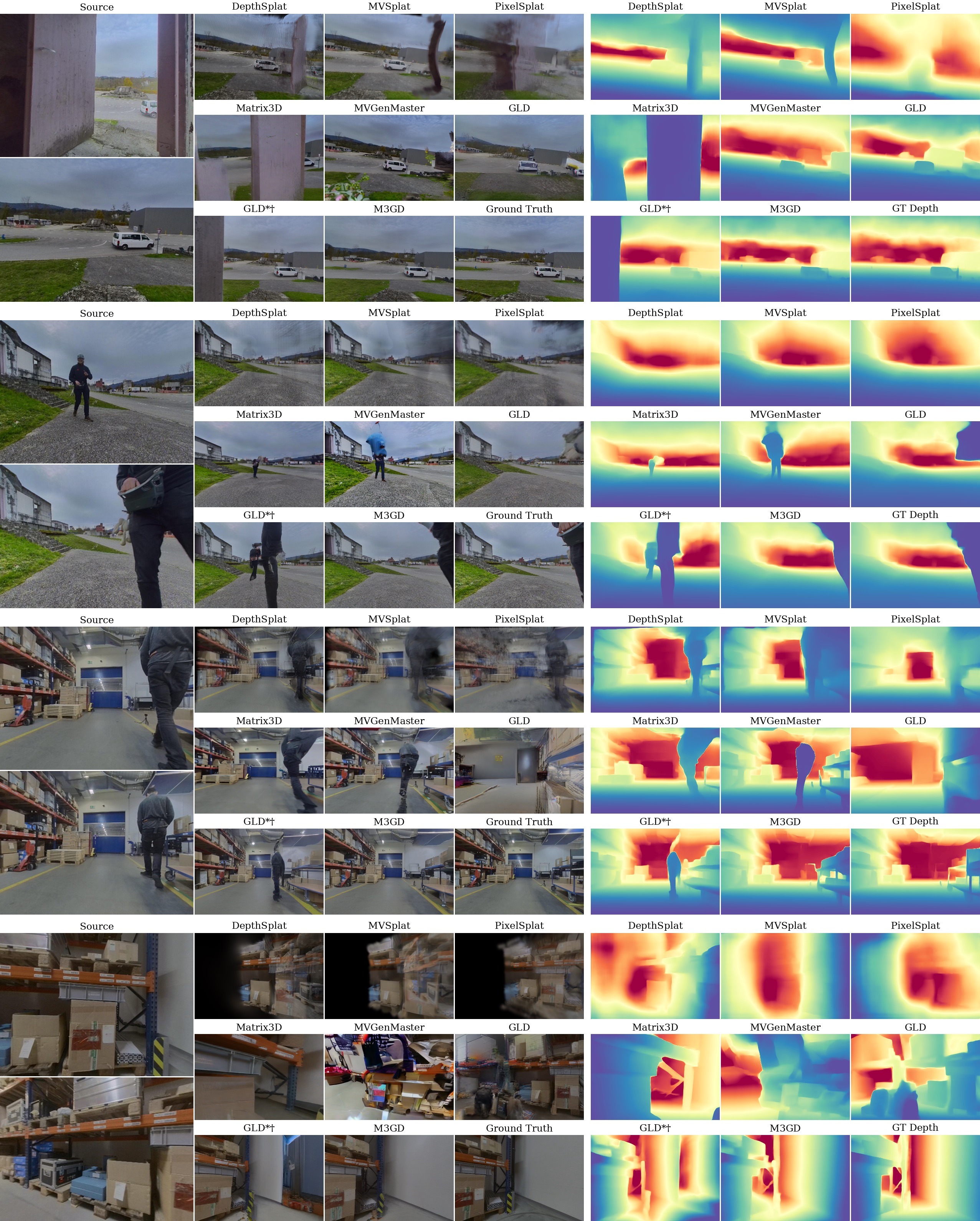}
    \vspace{-10pt}
    \caption{
    Qualitative RGB and depth comparison on the GrandTour evaluation split.
    For each scene, the two source images are shown on the left.
    The middle block compares RGB predictions, and is organized into three rows: feed-forward splatting baselines DepthSplat~\cite{xu2025depthsplat}, MVSplat~\cite{chen2024mvsplat}, and PixelSplat~\cite{charatan2024pixelsplat}; diffusion-based baselines Matrix3D~\cite{lu2025matrix3d} and MVGenMaster~\cite{cao2025mvgenmaster} together with the original GLD model~\cite{jang2026gld}; and GLD$^{*\dagger}$, our GrandTour-fine-tuned image-only GLD baseline, followed by M3GD and the ground-truth target view.
    The right block shows the corresponding DA3 depth estimates for the same predicted views and the ground-truth target depth, using the same method layout.
    Compared with splatting and image-only generative baselines, M3GD better preserves target-view appearance and produces depth maps that more closely follow the ground-truth scene geometry in challenging robot-captured scenes.
}
    \vspace{-10pt}
    \label{fig:qual_all_methods}
\end{figure*}
\textbf{Qualitative comparison.}
Fig.~\ref{fig:qual_all_methods} visualizes both RGB predictions and their corresponding DA3 depth estimates.
Under sparse robot-view inputs, splatting methods often produce warped or ghosted renderings around nearby structures and moving objects, and these artifacts are reflected in distorted depth maps.
Diffusion-based baselines can synthesize plausible RGB images, but their depth estimates often reveal viewpoint hallucinations or geometry that is inconsistent with the requested target view.
The fine-tuned image-only GLD$^{*\dagger}$ baseline is substantially stronger, but it remains limited to appearance and camera conditioning.
By using the target-view LiDAR packet as a geometric query, M3GD better anchors the target-view layout when the source images are ambiguous or do not reveal the target-facing surface.
This effect is visible in both the RGB predictions and the DA3 depth visualizations, consistent with the larger depth gains in Table~\ref{tab:baseline} and the LiDAR conditioning ablations in Table~\ref{tab:ablation_lidar}.

\begin{figure*}[!t]
    \centering
    \includegraphics[width=\textwidth]{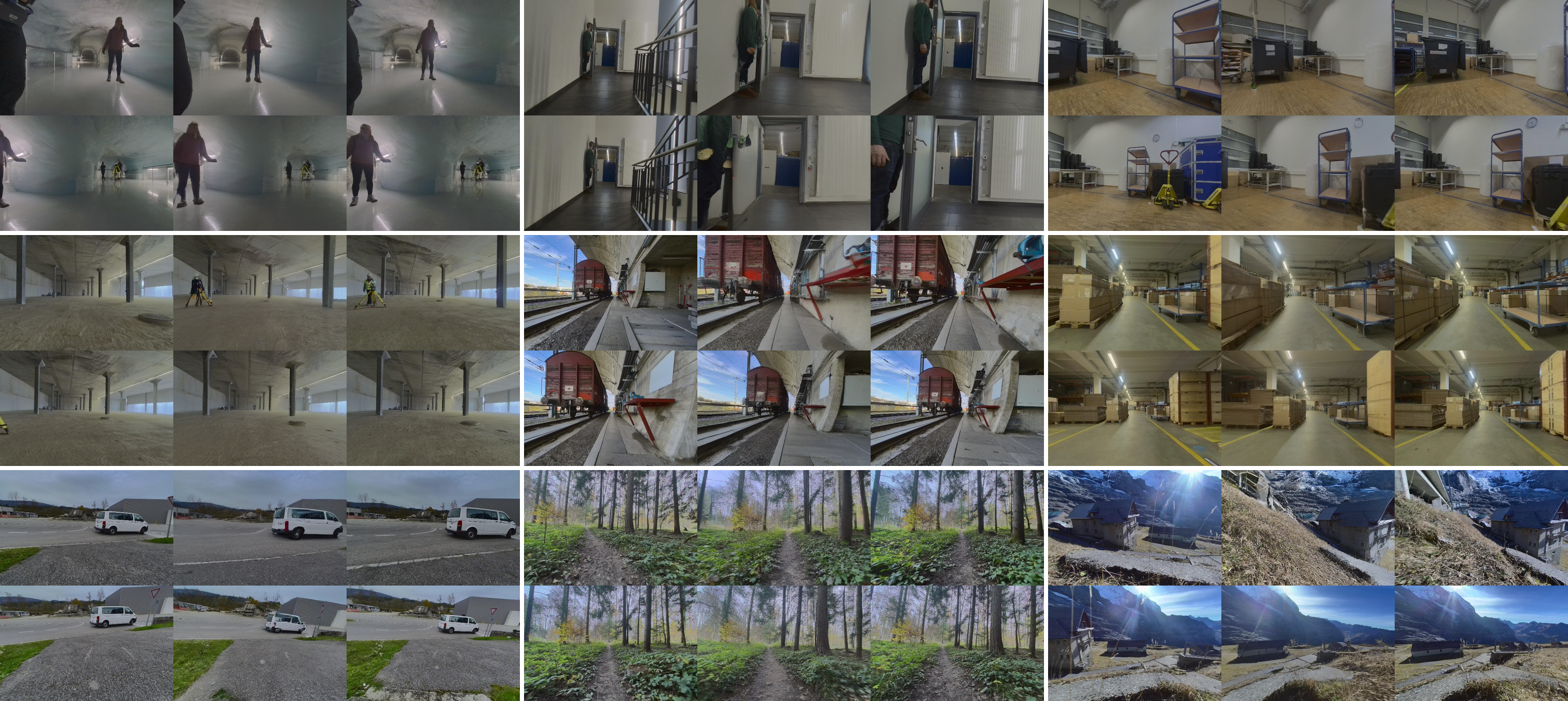}\\[-0.3em]
        {\scriptsize
    \makebox[\textwidth][c]{
        \makebox[0.1111\textwidth][c]{\rule[0.6ex]{0.0707\textwidth}{0.35pt}}
        \makebox[0.1111\textwidth][c]{\rule[0.6ex]{0.0707\textwidth}{0.35pt}}
        \makebox[0.1111\textwidth][c]{\rule[0.6ex]{0.0707\textwidth}{0.35pt}}
        \makebox[0.1111\textwidth][c]{\rule[0.6ex]{0.0707\textwidth}{0.35pt}}
        \makebox[0.1111\textwidth][c]{\rule[0.6ex]{0.0707\textwidth}{0.35pt}}
        \makebox[0.1111\textwidth][c]{\rule[0.6ex]{0.0707\textwidth}{0.35pt}}
        \makebox[0.1111\textwidth][c]{\rule[0.6ex]{0.0707\textwidth}{0.35pt}}
        \makebox[0.1111\textwidth][c]{\rule[0.6ex]{0.0707\textwidth}{0.35pt}}
        \makebox[0.1111\textwidth][c]{\rule[0.6ex]{0.0707\textwidth}{0.35pt}}
    }\\[-0.75em]
    \makebox[\textwidth][c]{
        \makebox[0.1111\textwidth][c]{Source}
        \makebox[0.1111\textwidth][c]{Prediction}
        \makebox[0.1111\textwidth][c]{Ground Truth}
        \makebox[0.1111\textwidth][c]{Source}
        \makebox[0.1111\textwidth][c]{Prediction}
        \makebox[0.1111\textwidth][c]{Ground Truth}
        \makebox[0.1111\textwidth][c]{Source}
        \makebox[0.1111\textwidth][c]{Prediction}
        \makebox[0.1111\textwidth][c]{Ground Truth}
    }}
    \vspace{-10pt}
    \caption{
    Qualitative visualization of M3GD on GrandTour evaluation samples.
    Each group follows the actual evaluation layout: two observed source views, two M3GD target-view predictions, and the corresponding ground-truth target views.
    M3GD synthesizes targets from sparse source images and target-view LiDAR geometry; target RGB is never provided.
    Examples are ordered from indoor scenes to semi-open structures and outdoor environments, showing that M3GD preserves dominant scene layout and reconstructs plausible target-view appearance across diverse and geometrically challenging robot-captured environments.
    }
    \label{fig:qual_ours_sequence}
    \vspace{-10pt}
\end{figure*}

\begin{table}
\centering
\caption{Effect of LiDAR conditioning across training modes (multi-step, level-$0$ latent). \textbf{Best} per column. {\footnotesize $^{\ddagger}$\,Off-the-shelf: frozen released GLD checkpoint, no LiDAR adapter. $^{\S}$\,As $^{\ddagger}$, with the LiDAR adapter trained. \emph{tgt-only} scores the loss on the target-view latents only, not on the jointly denoised source views. $^{\P}$\,Classifier-free LiDAR (cfg-dropout $0.1$); at evaluation a single CFG scale of $1.5$ is applied to the joint camera+LiDAR condition.}}
\label{tab:ablation_lidar}
\footnotesize
\setlength{\tabcolsep}{6pt}
\renewcommand{\arraystretch}{1.2}
\begin{tabular}{l c ccc}
\toprule
Training mode & LiDAR & PSNR\,$\uparrow$ & SSIM\,$\uparrow$ & LPIPS\,$\downarrow$ \\
\midrule
\multirow{2}{*}{freeze} & \ding{55}$^{\ddagger}$ & $17.39$ & $0.380$ & $0.468$ \\
                        & \ding{51}$^{\S}$       & $19.11$ & $0.434$ & $0.394$ \\
\midrule
\multirow{2}{*}{scratch}  & \ding{55} & $21.53$ & $0.520$ & $0.311$ \\
                          & \ding{51} & $22.16$ & $0.534$ & $0.295$ \\
\midrule
\multirow{2}{*}{fine-tune} & \ding{55} & $21.56$ & $0.525$ & $0.303$ \\
                          & \ding{51} & $22.22$ & $0.542$ & $0.285$ \\
\midrule
\multirow{3}{*}{\shortstack[l]{fine-tune\\(tgt-only loss)}} & \ding{55} & $21.62$ & $0.531$ & $0.302$ \\
                          & \ding{51}            & $22.27$ & $0.548$ & $0.287$ \\
                          & \ding{51}\,(cfg)$^{\P}$ & $\mathbf{22.33}$ & $\mathbf{0.549}$ & $\mathbf{0.283}$ \\
\bottomrule
\end{tabular}
\vspace{-10pt}
\end{table}
\subsection{Mechanism Ablations}\label{subsec:ablation}
Our mechanism ablations examine three questions: whether LiDAR conditioning helps across training modes (Section~\ref{subsubsec:lidar-training-mode}), whether the gain requires real per-sample LiDAR content rather than merely the added conditioning pathway, and which part of the packet carries the signal (both in Section~\ref{subsubsec:lidar-input-mode}). We keep these ablations separate from deployment-recipe choices (Section~\ref{subsec:deployment-recipe-ablations}).

\subsubsection{LiDAR Conditioning and Training Mode}\label{subsubsec:lidar-training-mode}
Table~\ref{tab:ablation_lidar} reports appearance quality as we vary the training mode and LiDAR conditioning. The freeze/no-LiDAR row is therefore not directly comparable to the GLD row of Table~\ref{tab:baseline} ($17.39$ vs.\ $17.92$), which runs GLD's native two-stage cascade. Three findings stand out. First, LiDAR conditioning improves every metric in every training mode, with per-mode PSNR gains ranging from $+0.63$ to $+1.72$\,dB; projected LiDAR geometry is thus a consistently useful signal for synthesis. Second, the gain is largest under the freeze setting ($+1.72$\,dB PSNR, $+0.054$ SSIM, $-0.074$ LPIPS), where the backbone is fixed and only the LiDAR adapter is trainable. Because the frozen backbone cannot adapt its appearance representation, this improvement is attributable to the LiDAR signal itself rather than to added model capacity, cleanly isolating LiDAR's contribution. Third, LiDAR does not substitute for backbone adaptation: although it lifts the frozen model substantially ($17.39\!\rightarrow\!19.11$\,PSNR), the absolute quality stays well below the trained variants, so the strongest results still require updating the backbone. Among the trained modes, fine-tuning from the pretrained GLD checkpoint slightly outperforms training from scratch ($22.22$ vs.\ $22.16$\,PSNR with LiDAR), and scoring the loss on the target views only, rather than on all jointly denoised views (\emph{tgt-only}), gives a further improvement ($22.27$). Training the LiDAR branch with classifier-free dropout adds a final increment, yielding our best configuration at $22.33\,/\,0.549\,/\,0.283$ (PSNR\,/\,SSIM\,/\,LPIPS). We therefore adopt fine-tuning with a target-only loss and classifier-free LiDAR conditioning as our default recipe.

\subsubsection{LiDAR Input Mode}\label{subsubsec:lidar-input-mode}
\begin{table}[t]
\centering
\caption{Effect of LiDAR input mode on the \emph{hard subset} (maximum camera baseline between views ${>}1$\,m, $n{=}1{,}407$; Section~\ref{subsec:dataset-metrics}). Each row is a separately trained model under the default recipe of Section~\ref{subsubsec:lidar-training-mode}; only the LiDAR input changes. Highlighting follows Table~\ref{tab:baseline}.}
\label{tab:lidar_input_mode}
\footnotesize
\setlength{\tabcolsep}{6pt}
\renewcommand{\arraystretch}{1.2}
\begin{tabular}{l ccc}
\toprule
LiDAR input mode & PSNR\,$\uparrow$ & SSIM\,$\uparrow$ & LPIPS\,$\downarrow$ \\
\midrule
Full (real packet)        & $\mathbf{21.00}$ & $\mathbf{0.476}$ & $\mathbf{0.314}$ \\
Geom-only                 & \underline{$20.90$} & \underline{$0.471$} & \underline{$0.319$} \\
Zero                      & $20.37$ & $0.455$ & $0.336$ \\
Learnable param           & $20.37$ & $0.455$ & $0.336$ \\
\midrule
\textit{No LiDAR (floor)} & $20.31$ & $0.455$ & $0.336$ \\
\bottomrule
\end{tabular}
\vspace{-10pt}
\end{table}
\begin{figure*}[!t]
    \centering
    \scriptsize
    \makebox[0.125\textwidth][c]{Source $0$}
    \makebox[0.125\textwidth][c]{Source $1$}
    \makebox[0.125\textwidth][c]{GLD$^{*\dagger}$~\cite{jang2026gld}}
    \makebox[0.125\textwidth][c]{Raw LiDAR}
    \makebox[0.125\textwidth][c]{Downsampled LiDAR}
    \makebox[0.125\textwidth][c]{Geom-only}
    \makebox[0.125\textwidth][c]{M3GD}
    \makebox[0.125\textwidth][c]{Ground Truth}

    \vspace{0.2em}
    \includegraphics[width=\textwidth]{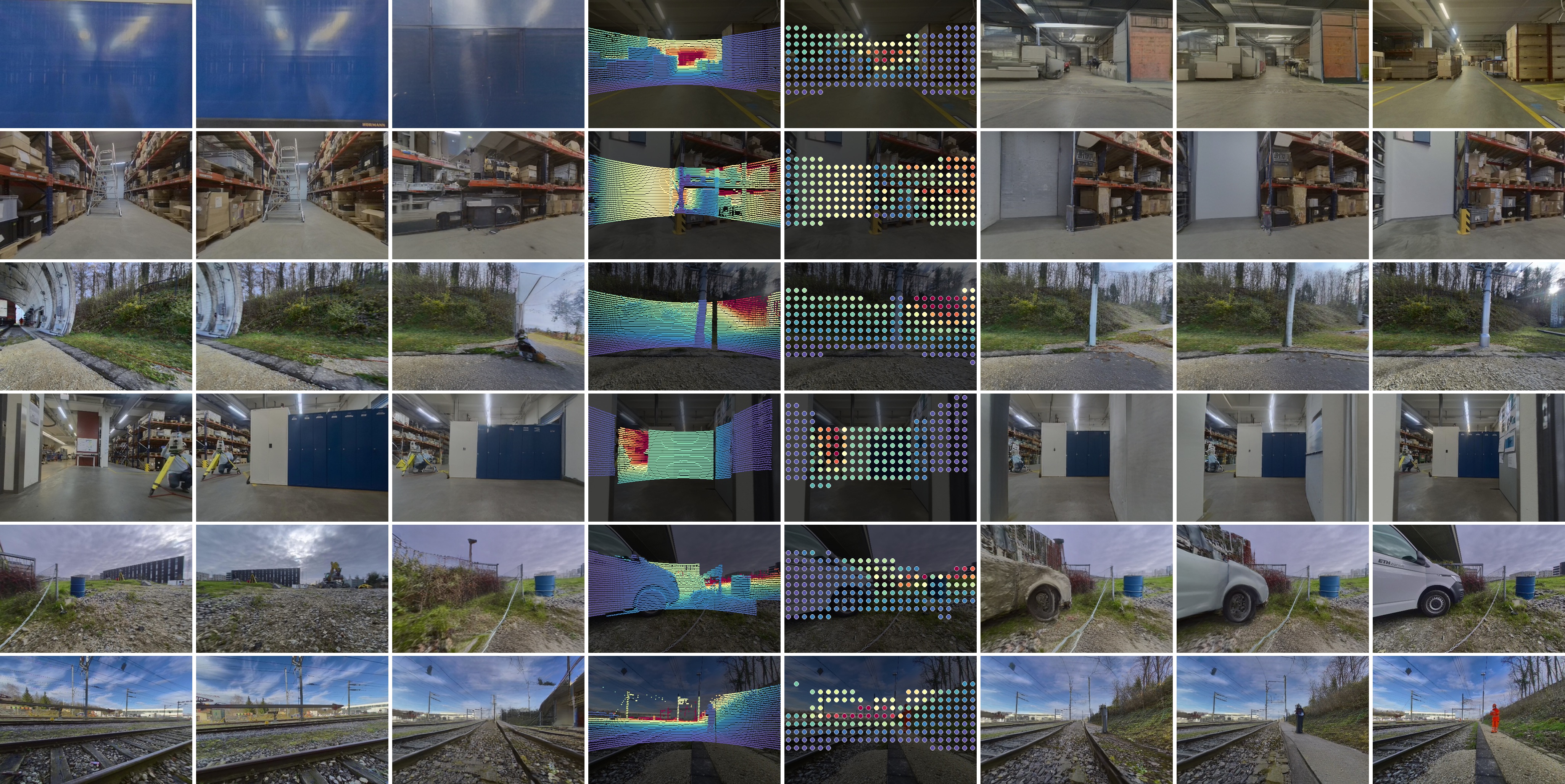}
    \vspace{-10pt}
    \caption{
    Qualitative examples highlighting the effect of target-view LiDAR conditioning.
    Each row shows two source images, the image-only GLD$^{*\dagger}$ baseline, the raw target-view LiDAR projection, our downsampled LiDAR representation, the geometry-only (geom-only) variant, M3GD, and the ground-truth target image.
    These examples contain large source-target viewpoint changes where RGB-only conditioning is ambiguous and GLD$^{*\dagger}$ often fails to infer the target-facing surface.
    The LiDAR-conditioned geom-only model recovers the dominant scene layout more accurately, while M3GD further improves appearance fidelity and local details such as boxes, road surfaces, walls, and vehicles by combining LiDAR geometry with Utonia features.
    }
    \label{fig:qual_lidar_ablation}
    \vspace{-10pt}
\end{figure*}
Table~\ref{tab:lidar_input_mode} separates the effect of the LiDAR packet \emph{content} from the effect of the added conditioning pathway. To sharpen the comparison, this ablation is evaluated on the \emph{hard subset} (Section~\ref{subsec:dataset-metrics}): a large camera baseline implies substantial ego-motion within the clip, which is precisely the regime where projected LiDAR conditioning has room to act. Each row is a separately trained model under the default recipe (multi-step, target-only fine-tune, classifier-free LiDAR). Only the LiDAR input supplied during training and evaluation changes. Two controls remove per-sample content at different strengths: \emph{Zero} feeds all-zero packets, so the LiDAR pathway receives no signal and effectively never trains, whereas \emph{Learnable param} replaces the packet with a shared, data-independent learned tensor of the same shape---a capacity-matched control whose input pathway does train but which carries no per-sample information. \emph{Geom-only} zeroes the learned 3D foundation feature (Utonia), isolating the hand-crafted geometric statistics.

\begin{figure*}[!t]
    \centering
    {\scriptsize
    \makebox[0.07\textwidth][c]{}%
    \makebox[\dimexpr0.93\textwidth/7\relax][c]{Source $0$}%
    \makebox[\dimexpr0.93\textwidth/7\relax][c]{Source $1$}%
    \makebox[\dimexpr0.93\textwidth/7\relax][c]{Ground Truth}%
    \makebox[\dimexpr0.93\textwidth/7\relax][c]{Our RGB}%
    \makebox[\dimexpr0.93\textwidth/7\relax][c]{\shortstack{Our Depth}}%
    \makebox[\dimexpr0.93\textwidth/7\relax][c]{\shortstack{GLD RGB}}%
    \makebox[\dimexpr0.93\textwidth/7\relax][c]{\shortstack{GLD Depth}}
    }\\[0.08em]
    \parbox[b][0.087\textwidth][c]{0.07\textwidth}{\centering\scriptsize\shortstack{Q1-RGB\\T1}}%
    \includegraphics[width=0.93\textwidth]{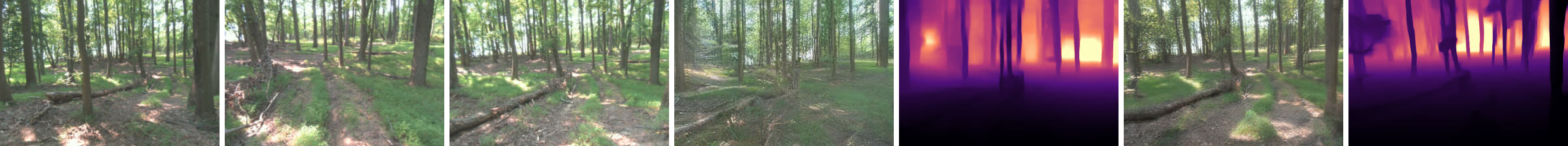}\\[0.08em]
    \parbox[b][0.087\textwidth][c]{0.07\textwidth}{\centering\scriptsize\shortstack{Q1-LiDAR\\T2}}%
    \includegraphics[width=0.93\textwidth]{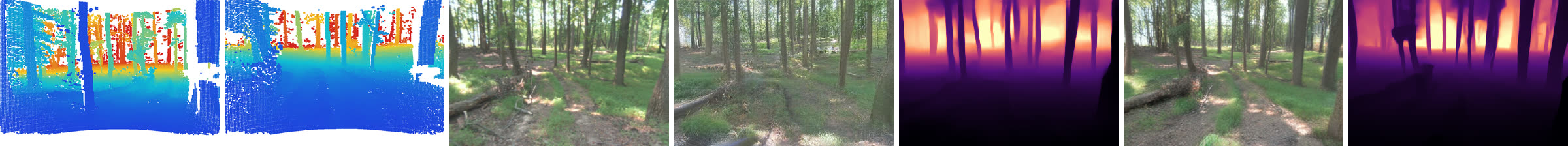}\\[0.08em]
    \parbox[b][0.087\textwidth][c]{0.07\textwidth}{\centering\scriptsize\shortstack{Q2-RGB\\T1}}%
    \includegraphics[width=0.93\textwidth]{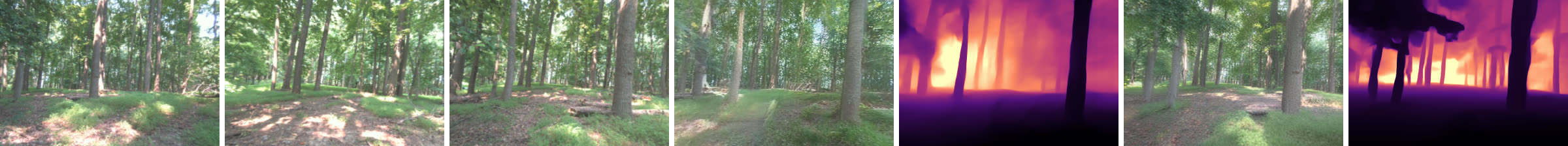}\\[0.08em]
    \parbox[b][0.087\textwidth][c]{0.07\textwidth}{\centering\scriptsize\shortstack{Q2-LiDAR\\T2}}%
    \includegraphics[width=0.93\textwidth]{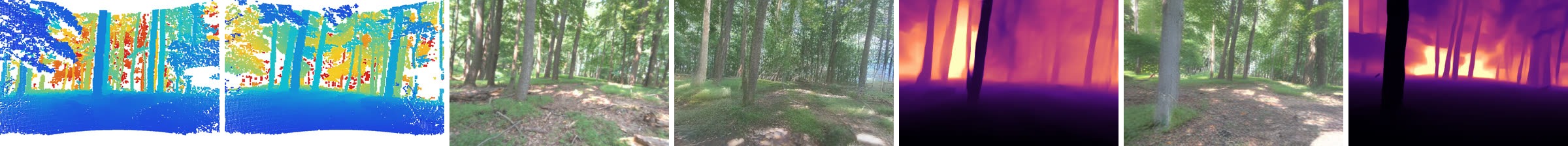}\\%
    \vspace{-10pt}
    \caption{
    Qualitative results for two real-robot queries Q1 (rows $1$--$2$) and Q2 (rows $3$--$4$), using $K=10$ Euler steps, $N=2$ source views, and $M=2$ target views.
    Each query forms a two-row block: in the first two columns, the upper row shows the two source RGB images and the lower row shows their corresponding projected LiDAR point clouds.
    The remaining columns show the ground-truth RGB, M3GD synthesized RGB, our RGB readout depth, GLD synthesized RGB, and GLD RGB readout depth (defined in Section~\ref{subsec:baseline}) for target T1 in the upper row and target T2 in the lower row.
    }
    \label{fig:real-robot-samples}
    \vspace{-10pt}
\end{figure*}

Both controls land essentially at the no-LiDAR floor: $20.37$ for both Zero and Learnable param vs.\ $20.31$ PSNR, with SSIM and LPIPS matching the floor exactly. The $+0.69$\,dB gain of the full packet ($21.00$ vs.\ $20.31$) is therefore driven by real, per-sample LiDAR content, not by extra parameters, extra input channels, or a learned input bias, which the capacity-matched control rules out. Geom-only ($20.90$) recovers most of the full-packet gain, indicating that geometric statistics alone preserve the bulk of the improvement in this subset, with the learned feature contributing a small additional margin.

\textbf{Generalization across scene types.}
Fig.~\ref{fig:qual_ours_sequence} shows M3GD predictions with examples ordered from indoor scenes through semi-open structures to outdoor environments. Across this range, the synthesized views preserve dominant geometric structures---walls, ground planes, rails, shelves, corridors, vehicles, and vegetation---while inferring plausible appearance from only two source images. This behavior is consistent with the intended role of the target-view LiDAR conditioning packet as a geometric anchor: the predicted layout remains stable even where the sparse and sometimes ambiguous source observations constrain appearance only weakly. Remaining artifacts concentrate around fine structures, reflective regions, and dynamic objects, precisely where that constraint is weakest.
Nevertheless, the synthesized target views remain geometrically coherent in challenging indoor, semi-open, and outdoor robot-captured scenes, supporting the quantitative improvements reported in Table~\ref{tab:baseline}.

\textbf{Effect of LiDAR conditioning.}
Fig.~\ref{fig:qual_lidar_ablation} isolates the contribution of target-view LiDAR conditioning on examples with large source--target viewpoint changes, in which target-facing walls, road regions, vehicles, shelves, and open corridors are only weakly visible, or invisible, in the source views. In this regime, the image-only GLD$^{*\dagger}$ baseline often synthesizes plausible but incorrectly placed content, reflecting the ambiguity of inferring the target view from RGB observations alone. The LiDAR columns of the figure show the signal that resolves this ambiguity: the raw projection of the target-view sweep carries the geometry available at the queried pose, and its downsampled form on the image-latent grid is the compact version the packet delivers to the model. Conditioned on this signal, the geom-only variant already recovers the dominant target-view structure more accurately than the image-only baseline, confirming that the explicit geometry statistics act as a geometric anchor. Geometry alone, however, does not resolve appearance: geom-only outputs can still miss texture, material detail, and fine object appearance. The full packet closes this gap by adding the downsampled Utonia descriptors, yielding sharper local details such as boxes, walls, road surfaces, vegetation, and vehicles.
This qualitative behavior is consistent with the LiDAR conditioning ablations in Table~\ref{tab:lidar_input_mode}.

\subsection{Deployment Recipe Ablations}\label{subsec:deployment-recipe-ablations}

After the ablations isolate why the RGB--LiDAR representation helps, we choose the practical operating point for robot deployment: the Euler integration-step budget $K$ and the number $M$ of jointly synthesized target views. These recipe choices are not the main scientific claim, they fix the quality--cost trade-off under the deployment budget. The  level-$0$ design itself is a stated choice of Section~\ref{sec:method}, motivated by the alignment result and the deployment budget and Table~\ref{tab:deployment_scaling} supports it from the cost side. The resulting quality, latency, and memory scaling across integration-step and target-view counts are reported in Section~\ref{subsec:deployment-runtime}.

\subsection{Real-World Robot Deployment}\label{subsec:real-robot-deployment}

The ablations above isolate the mechanism on held-out recordings. We close by exercising the full conditioning pathway in the loop on a physical robot, in exactly the regime Section~\ref{subsec:problem-setup} describes. The Warthog platform of Section~\ref{subsec:robot-platform} (Fig.~\ref{fig:warthog-platform}) drives a route while logging its camera and LiDAR streams. To synthesize the view at a chosen pose along the traversal, we take a short window of past, time-aligned camera--LiDAR pairs as conditioning views, project the LiDAR sweep temporally nearest the target query's timestamp into the target view to form the target packet (Section~\ref{subsec:projected-rgb-lidar-representation}, no target RGB is used), and predict the target-view RGB and depth with the same flow model used in the aforementioned experiments. We emphasize that this model is deployed as is: it is trained on GrandTour only and is neither retrained nor fine-tuned on any data from the Warthog, so the deployment is a zero-shot transfer to a new robot platform, sensor suite, and environment. The representative outputs shown below use $K{=}10$ Euler integration steps. Every input the model needs is a byproduct of the robot's normal operation, available at the moment it needs the synthesized view.

The two selected queries, Q1 and Q2, lie on a wooded portion of the route about half a minute apart; each predicts RGB and depth at two held-out poses, T1 and T2, logged two seconds apart, from two conditioning views recorded within a few meters of them. Unlike the benchmark recordings, these inputs contain deployment artifacts visible in the projected point clouds, including LiDAR shadows caused by viewpoint occulsion and residual camera-LiDAR misalignment due to imperfect temporal synchronization. Fig.~\ref{fig:real-robot-samples} shows the outputs. Nevertheless, they reproduce the behavior observed on held-out GrandTour recordings: the synthesized views follow the requested camera poses, and the decoded depth preserves the dominant layout supported by the projected LiDAR, including the ground plane, tree trunks, and large background structure. The point of this demonstration is not a new aggregate metric or controlled robustness benchmark. It is evidence that the conditioning signal required by M3GD is available from the synchronized camera--LiDAR stream of a physical robot, remains useful in the presence of ordinary sensing artifacts, and transfers to a previously unseen platform and sensor configuration without any target-domain retraining.

\begin{table*}
    \centering
    \setlength{\tabcolsep}{6pt}
    \caption{
        On-robot quality, latency, and memory scaling on the Warthog platform with two fixed source views  ($N=2$). 
        Quality is averaged over $30$ queries; latency and peak GPU reserved memory are measured over four synchronized post-warm-up queries.
    }
    \label{tab:deployment_scaling}

    \footnotesize

    \begin{tabular}{cc|ccc|ccc}
        \toprule
        \multicolumn{2}{c}{Configuration}
        & \multicolumn{3}{c}{M3GD Quality}
        & \multicolumn{3}{c}{M3GD Computational Cost} \\
        \cmidrule(lr){1-2}
        \cmidrule(lr){3-5}
        \cmidrule(lr){6-8}

        \shortstack{Steps $K$}
        & \shortstack{Targets $M$}
        & \shortstack{PSNR (dB) $\uparrow$}
        & \shortstack{SSIM $\uparrow$}
        & \shortstack{LPIPS $\downarrow$}
        & \shortstack{Latency (s) $\downarrow$}
        & \shortstack{Lat./Target (s) $\downarrow$}
        & \shortstack{Mem. (GiB) $\downarrow$} \\
        \midrule

        $1$  & $2$ & $14.13$ & $0.279$ & $0.889$ & $\mathbf{3.13 \pm 0.44}$ & $1.56 \pm 0.22$ & $\mathbf{4.22}$ \\
        
        \midrule

        $5$  & $2$ & $\mathbf{15.86}$ & $\mathbf{0.281}$ & $0.480$ & $\underline{3.63 \pm 0.45}$ & $1.81 \pm 0.22$ & $\underline{5.70}$ \\
        $5$  & $3$ & $15.65$ & $\underline{0.280}$ & $0.489$ & $3.77 \pm 0.45$ & $1.26 \pm 0.15$ & $5.96$ \\
        $5$  & $4$ & $15.54$ & $0.276$ & $0.494$ & $3.91 \pm 0.44$ & $\mathbf{0.98 \pm 0.11}$ & $6.11$ \\

        \midrule

        $10$ & $2$ & $\underline{15.84}$ & $0.270$ & $\mathbf{0.476}$ & $4.26 \pm 0.44$ & $2.13 \pm 0.22$ & $5.80$ \\
        $10$ & $3$ & $15.67$ & $0.270$ & $0.481$ & $4.54 \pm 0.45$ & $1.51 \pm 0.15$ & $6.05$ \\
        $10$ & $4$ & $15.58$ & $0.267$ & $0.483$ & $4.83 \pm 0.44$ & $\underline{1.21 \pm 0.11}$ & $6.25$ \\

        \midrule

        $20$ & $2$ & $15.74$ & $0.262$ & $\underline{0.478}$ & $5.50 \pm 0.44$ & $2.75 \pm 0.22$ & $5.97$ \\
        $20$ & $3$ & $15.59$ & $0.261$ & $0.481$ & $6.08 \pm 0.45$ & $2.03 \pm 0.15$ & $6.29$ \\
        $20$ & $4$ & $15.51$ & $0.259$ & $0.481$ & $6.69 \pm 0.44$ & $1.67 \pm 0.11$ & $6.50$ \\

        \bottomrule
    \end{tabular}
\end{table*}
\subsection{Onboard Runtime Tradeoff}\label{subsec:deployment-runtime}
Table~\ref{tab:gld-vs-m3gd} establishes that M3GD's single-pass design is more efficient than GLD's native two-stage cascade. We next characterize deployment runtime and GPU memory utilization on our robot setup across an out-of-distribution route, varying the number of Euler integration steps $K\in\{1,5,10,20\}$ and the number of jointly synthesized target views $M\in\{2,3,4\}$. Changing $K$ modifies only the inference solver, and every operating point uses the same trained weights. We evaluate $K{=}1$ at $M{=}2$ and the full $M\in\{2,3,4\}$ sweep for $K\in\{5,10,20\}$, yielding the $10$ operating points per runtime branch reported in Table~\ref{tab:deployment_scaling}. All points hold $N=2$ fixed, and source observations and shared target frames do not change as $M$ increases. Hardware and the timing protocol are described in Section~\ref{subsec:robot-platform}.

For $K=10$, $N=2$, $M=2$, M3GD peaks at $5.80$~GiB and spans $4.22$--$6.50$~GiB across the sweep. The step sweep demonstrates an inference-budget trade-off, where a single Euler step gives the lowest M3GD latency ($3.13$~s for two targets), but its LPIPS degrades to $0.889$. Five steps raise PSNR to $15.86$~dB and SSIM to $0.281$, reduce LPIPS to $0.480$, and require $3.63$~s, while additional steps increase latency without improving the appearance metrics on this route. Jointly synthesizing more targets increases total latency and memory, but amortizes computation: for M3GD at $K{=}10$, latency per target falls from $2.13$~s at $M{=}2$ to $1.21$~s at $M{=}4$. Thus when a robot requires multiple queried views from the same source context, M3GD can reuse shared computation, such as LiDAR encoding, across targets rather than paying a full generation cost for each view.

Together, Tables~\ref{tab:gld-vs-m3gd} and~\ref{tab:deployment_scaling} support complementary claims. Table~\ref{tab:gld-vs-m3gd} shows that LiDAR conditioning does not require a heavier diffusion architecture: M3GD achieves its evaluation gains while avoiding GLD's multi-stage cascade. Notably, LiDAR conditioning shifts computational cost upfront and yields substantial savings when predicting multiple views. Table~\ref{tab:deployment_scaling} shows that the same M3GD design provides selectable quality, latency, and memory operating points.

\section{Limitations and Future Works}\label{sec:limitations}
\textbf{Temporal staleness.} M3GD selects the LiDAR sweep nearest each image timestamp, but the current evaluation does not isolate the effect of camera--LiDAR delay. A controlled time-offset ablation and explicit modeling of timing uncertainty inside the generative network can be interesting future directions to explore.

\textbf{Per-cell averaging mixes surfaces.} Pooling all points that project into a latent cell averages across depth layers near object boundaries and thin structures. Front-surface-only pooling, depth-sorted visibility weighting, or multi-layer packets are natural refinements. We deliberately kept the residual adapter minimal to make the ablations attributable.

\textbf{Scale and scope of evaluation.} Both the NVS benchmark and the representation study use GrandTour recordings from one robot platform (ANYmal-D). Extending the evaluation to wide-baseline cross-camera synthesis, other platforms, and driving datasets would probe how far the measured DA3--Utonia structure transfers across domains; the packet interface itself is platform-agnostic. We will release our code and evaluation protocols to facilitate these extensions. The comparison against feed-forward splatting baselines is matched in fine-tuning data but cannot perfectly equalize pretraining corpora across methods.

\textbf{Dynamic scenes.} Packets assume the time-nearest sweep describes the scene at the image timestamp. Fast-moving objects violate this even at small $\Delta t$; neither the packet schema (beyond the variance channel) nor the model addresses object motion explicitly.

\textbf{Inference budget and deployment hardware.} Reducing the Euler integration budget lowers latency, but single-step sampling substantially degrades LPIPS on the deployment route. The reported runtimes were measured on a robot equipped with an RTX A6000 as described in Section~\ref{subsec:robot-platform}, so they should not be interpreted as guaranteed real-time performance on smaller onboard computers. Adaptive solvers and accelerated flow sampling are promising directions for improving this quality--cost trade-off. 

\section{Conclusion}\label{sec:conclusion}
We presented M3GD, a robotic novel view synthesis approach that composes independently trained 2D image and 3D point-cloud foundation models through projection, without pretraining a cross-modal translator. The key observation is that, after camera projection, frozen point-cloud features share substantial spatial structure with frozen image-foundation latents. This makes a simple RGB--LiDAR packet viable: the generative backbone keeps its latent space, decoders, camera conditioning, and training objective, while each view receives spatially aligned 3D evidence on the grid it already uses for synthesis.

On GrandTour, M3GD improves target-view RGB and depth synthesis over an image-only version of the same backbone, with the largest gains on depth metrics. Ablations show that the benefit comes from real, pixel-aligned LiDAR content rather than adapter capacity, and that the target-view packet acts as a geometric query linking the requested view to the source observations without revealing target appearance. On a ground robot with a different sensor suite, the same GrandTour-trained model runs without retraining, and the number of Euler steps trades latency for perceptual quality. Together, these results suggest a practical pattern for robotic perception: once 2D and 3D foundation representations are projected into a shared view-local frame, a lightweight conditioning pathway can expose their complementary knowledge to a generative model without rebuilding the synthesizer around a new modality.
 
\bibliographystyle{IEEEtran}
\bibliography{mybib}

\end{document}